%% file: 0_main.tex
\documentclass[10pt,twocolumn,letterpaper]{article}
\pdfoutput=1
\usepackage[pagenumbers]{cvpr}
\usepackage{graphicx}
\usepackage{amsmath}
\usepackage{amssymb}
\usepackage{booktabs}
\usepackage{times}
\usepackage{graphicx}
\usepackage{caption}
\usepackage{subcaption}
\usepackage{siunitx}
\usepackage[usenames,dvipsnames]{xcolor}
\usepackage[pagebackref,breaklinks,colorlinks]{hyperref}

\DeclareMathOperator*{\argmin}{argmin}
\usepackage[accsupp]{axessibility}

\usepackage[capitalize]{cleveref}
\crefname{section}{Sec.}{Secs.}
\Crefname{section}{Section}{Sections}
\Crefname{table}{Table}{Tables}
\crefname{table}{Tab.}{Tabs.}

\begin{document}

\title{Towards Understanding Adversarial Robustness of Optical Flow Networks}

\author{Simon Schrodi \hspace{1cm} Tonmoy Saikia \hspace{1cm} Thomas Brox\\
University of Freiburg, Germany\\
{\tt\small \{schrodi, saikiat, brox\}@cs.uni-freiburg.de}
}
\maketitle

\begin{abstract}
Recent work demonstrated the lack of robustness of optical flow networks to physical patch-based adversarial attacks. The possibility to physically attack a basic component of automotive systems is a reason for serious concerns. In this paper, we analyze the cause of the problem and show that the lack of robustness is rooted in the classical aperture problem of optical flow estimation in combination with bad choices in the details of the network architecture. We show how these mistakes can be rectified in order to make optical flow networks robust to physical patch-based attacks. Additionally, we take a look at global white-box attacks in the scope of optical flow. We find that targeted white-box attacks can be crafted to bias flow estimation models towards any desired output, but this requires access to the input images and model weights. However, in the case of universal attacks, we find that optical flow networks are robust. Code is available at \url{https://github.com/lmb-freiburg/understanding_flow_robustness}.
\end{abstract}

\input{1_intro}
\input{2_related_work}
\input{3_methods}
\input{3_understanding}
\input{4_experiments}
\input{4b_global_attacks}
\input{5_conclusion}

\section*{Acknowledgements}
Funded by the Deutsche Forschungsgemeinschaft (DFG) -- BR 3815/10-1,
INST 39/1108-1, and the German Federal Ministry for Economic Affairs and Climate Action" within the project KI Delta Learning -- 19A19013N.

{\small
\bibliographystyle{ieee_fullname}
\bibliography{egbib}
}

\input{supplement}

\end{document}

%% file: 1_intro.tex
\section{Introduction}
While deep learning has been conquering many new application domains, it has become increasingly evident that deep networks are vulnerable to distribution shifts.
Adversarial attacks are a particular way to showcase this vulnerability,
where one finds the minimal input perturbation that is sufficient to corrupt the network output. As the small perturbation moves the sample out of the training distribution, the network is detached from its learned patterns and follows the suggestive pattern of the attack. Although many methods have been proposed to improve robustness~\cite{xu2020adversarial}, they only alleviate the problem but do not solve it~\cite{athalye2018robustness}.  

\begin{figure}[t]
    \centering
    \includegraphics[width=0.47\textwidth]{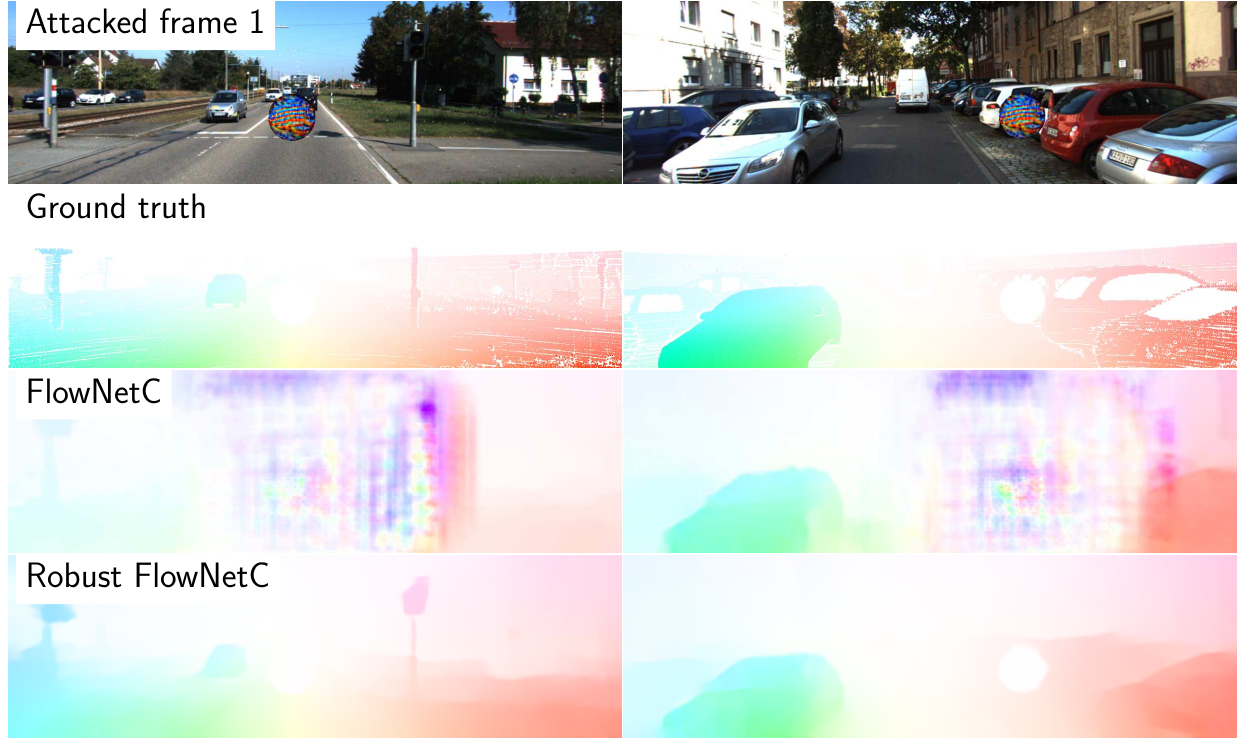}
    \caption{\textbf{Overview.} Physical patch-based adversarial attacks on optical flow can be avoided by minor architectural changes. First row: attacked first frame. Second row: ground truth optical flow. Third and fourth rows: the resulting optical flow estimates of FlowNetC~\cite{dosovitskiy2015flownet} and our proposed Robust FlowNetC. FlowNetC is strongly affected by the adversarial patch, whereas Robust FlowNetC is barely affected.  For the robust version we make simple design changes based on causes of the attack; see Section~\ref{sec:makeFlowRobust}.}
    \label{fig:teaser}
\end{figure}

While most white-box adversarial attacks are mainly of academic relevance as they reveal the weaknesses of deep networks \wrt out-of-distribution data, physical adversarial attacks have serious consequences for safe deployment. In physical attacks, the input is not perturbed artificially, but a confounding pattern is placed in the real world to derail the machine learning approach.

Most work on adversarial attacks has been concerned with recognition problems, and it looked for a while as if correspondence problems are not a good target for adversarial attacks. However, Ranjan~\etal~\cite{ranjan2019attacking} showed that they can successfully perform physical adversarial patch attacks on optical flow networks. They optimized an adversarial local patch that they can paste into both images, such that large errors appear in the estimated optical flow field even far away from the affected image location. 
They also showed that the same adversarial patch worked on all vulnerable architectures, and even demonstrated physical attacks, where the printed patch is physically added to a scene and derails the optical flow estimation.
Ranjan~\etal found that different network architectures show different levels of vulnerability, whereas conventional optical flow methods are not vulnerable at all. They hypothesized the cause for the vulnerability to be in the common encoder-decoder architecture of FlowNet~\cite{dosovitskiy2015flownet} and its derivatives but did not provide a conclusive analysis.

In this paper, we continue their work by a deeper analysis of the actual reason behind the vulnerability. In particular, we answer the following questions.\\
\textbf{(1)} What is the true cause of adversarial patch attacks?\\
\textbf{(2)} Knowing the cause, can the patch-based attack also be built without optimizing it for the particular network (zero query black-box attack)?\\
\textbf{(3)} Can the severe vulnerability be avoided by a specific design of the network architecture or by avoiding mistakes in such design? For an overview see Figure~\ref{fig:teaser}.

After answering these questions positively, we turn towards (global) adversarial perturbation attacks, \ie, attacks that modify the whole image.  We demonstrate that any target optical flow field can be generated; see Figure \ref{fig:targetedAttack}. On the other hand, we show that this attack strategy does not apply to universal ({input-agnostic}) attacks, \ie, global attacks on optical flow networks must exploit the structure of the input images. 
This is different from unprotected recognition networks, which are vulnerable to imperceptible universal attacks~\cite{moosavi2017universal, hendrik2017universal}.

%% file: 2_related_work.tex
\section{Related Work}
\textbf{Optical flow.}
For many decades, optical flow was estimated with approaches that minimize an energy function consisting of a matching cost and a term that penalizes deviation from smoothness~\cite{horn1981determining,Black:CVIU:1996,brox2004high,brox2010large}. 

Inspired by the success of CNNs on recognition tasks, Dosovitskiy \etal \cite{dosovitskiy2015flownet} introduced estimation of optical flow with a deep network, by training it end-to-end.
They proposed two network architectures -- FlowNetS and FlowNetC -- of which the first is a regular encoder-decoder architecture, whereas the second includes an additional correlation layer that explicitly computes a cost volume for feature correspondences between the two images -- like the correlation approaches from the very early days of optical flow estimation, but integrated into the surrounding of a deep network for feature learning and interpretation of the correlation output. The concept of these architectures has been picked up by many follow-up works that introduced, for instance, coarse-to-fine estimation~\cite{ranjan2017optical,sun2018pwc}, stacking~\cite{ilg2017flownet}, and multi-scale 4D all-pairs correlation volumes combined with the separate use of a context encoder as well as a recurrent unit for iterative refinement~\cite{teed2020raft}. Most of the architectures have an explicit correlation layer like the original FlowNetC.

\textbf{Adversarial attacks.}
The first works that brought up the issue of vulnerability of deep networks to adversarial examples were in the context of image classification and generated the examples by solving a box-constrained optimization problem \cite{szegedy2013intriguing} or by perturbing the input images with the gradient \wrt the input \cite{goodfellow2014explaining}.
Su~\etal~\cite{su2019one} showed that neural networks can be even attacked by just changing a single pixel.
Kurakin~\etal~\cite{kurakin2016adversarial} showed that adversarial attacks also work in the physical world by printing out adversarial examples.
Several follow-up works have confirmed this behavior in other contexts~\cite{brown2017adversarial,eykholt2018robust,athalye2018synthesizing}.
Hendrycks~\etal~\cite{hendrycks2019natural} showed that adversarial examples can even exist in natural, real-world images, which relates adversarial attacks to the more general issue of out-of-distribution samples. 

Works on adversarial attacks concentrated on various sorts of recognition tasks, \ie, tasks where the output depends directly on some feature representation of the input image, such as classification, semantic segmentation, single-view depth estimation, or image retrieval. 
Recently, Ranjan~\etal~\cite{ranjan2019attacking} showed that optical flow networks are also vulnerable to adversarial patch attacks and can also attack flow networks in a real-world setting. From their experimental evidence, they hypothesized that the encoder-decoder architecture is the main cause for the adversarial vulnerability, whereas spatial pyramid architectures, as well as classical optical flow approaches, are robust to patch-based attacks. Further, they showed that flow networks are not spatially invariant and the deconvolutional layers lead to an amplification of activations as well as checkerboard artifacts \cite{odena2016deconvolution}.
Recently, Wong~\etal~\cite{wong2021stereopagnosia} showed that imperceptible perturbations added to each pixel individually can significantly deteriorate the output of stereo networks. They used adversarial data augmentation to make stereo networks more robust. While stereo networks are vulnerable to image-specific attacks, they showed that perturbations do not transfer well to the next time step.

%% file: 3_methods.tex
\section{Adversarial Patch Attacks}
\textbf{Adversarial patch.} 
Ranjan~\etal~\cite{ranjan2019attacking} proposed attacking flow networks by pasting a patch $p$ of resolution $h\!\times\!w$ onto the image frames $(I_{t},I_{t+1})\in\mathcal{I}$ of resolution $H\times W$ at the same location, rotation, and scaling. 
To craft an adversarial patch for flow network $F$, they minimized the cosine similarity between the unattacked flow $(u,v)$ and the attacked one $(\Tilde{u},\Tilde{v})$.
More formally, they optimized
\begin{equation}
\label{eq:patchOpt}
    \hat{p} = \argmin\limits_{p}\mathbb{E}_{(I_{t},I_{t+1})\sim\mathcal{I},l\sim\mathcal{L},\delta\sim\mathcal{T}} \frac{(u,v)\cdot  (\Tilde{u},\Tilde{v})}{||(u,v)||\cdot ||(\Tilde{u},\Tilde{v})||},
\end{equation}
where they randomly sample the location $l\in\mathcal{L}$ and affine transformations $\delta\in\mathcal{T}$, \ie, rotation and scaling, to generalize better to a real-world setting.

\textbf{Vulnerability of existing optical flow methods.}
\begin{table}
    \centering
    \caption{\textbf{Patch attacks on different flow networks.} We show average unattacked and worst-case attacked End-Point-Error (EPE) on the KITTI 2015 training dataset (for details see Section~\ref{sec:causesForAttacks}). We only show results for larger patch sizes ($102\!\times\!102$ and $153\!\times\!153$), since smaller patches show simply a weaker effect~\cite{ranjan2019attacking}.}
    \resizebox{0.82\columnwidth}{!}{%
    \begin{tabular}{l|S[table-format=3.2]|*{2}{S[table-format=3.2]}}
                  & {Un-}  & \multicolumn{2}{c}{Attacked} \\
          Network & {attacked}         & {\text{102x102}}  & {\text{153x153}}  \\
                  & {EPE}            & {\text{(2.1\%)}}  & {\text{(5.8\%)}}  \\
          \midrule
          FlowNetC~\cite{dosovitskiy2015flownet} & 11.50 & 52.66 & 51.99 \\
          FlowNetS~\cite{dosovitskiy2015flownet} & 14.33 & 17.35 & 17.92 \\
          FlowNet2~\cite{ilg2017flownet} & 10.07 & 12.40 & 13.36 \\
          SPyNet~\cite{ranjan2017optical} & 24.26 & 27.47 & 25.84 \\
          PWC-Net~\cite{sun2018pwc} & 12.55 & 18.08 & 17.70 \\
          RAFT~\cite{teed2020raft} & 5.86 & 8.48 & 9.01 \\
    \end{tabular}
    }
    \label{tab:additionalResults}
\end{table}
Ranjan~\etal~\cite{ranjan2019attacking} found that different flow network architectures show different degrees of vulnerability. Table~\ref{tab:additionalResults} shows the performance degradation of different architectures \wrt patch-based attacks.
FlowNetC is the only truly vulnerable flow network, whereas the others are much more robust.

Ranjan \etal~\cite{ranjan2019attacking} attributed the vulnerability to the encoder-decoder architecture and the higher robustness to the spatial pyramid of PWC-Net and SPyNet. However, there is a counterexample that proves this hypothesis wrong: FlowNetS -- the direct counterpart of FlowNetC \emph{without} correlation layer -- is a plain encoder-decoder architecture without a spatial pyramid and, as Table~\ref{tab:additionalResults} reveals, is quite robust to the attack. 
Thus, the encoder-decoder architecture cannot be the root cause for the vulnerability, even though the decoder can be responsible for amplifying the effect.\footnote{Like strong rain is the root cause for flooding but a dam (\ie, spatial pyramid) can avoid the flooding despite strong rain to a certain degree.}

%% file: 3_understanding.tex
\section{What Causes a Successful Patch Attack?}\label{sec:causesForAttacks}
We build on the attack procedure of Ranjan~\etal~\cite{ranjan2019attacking}, \ie, we also use KITTI 2012~\cite{Geiger2012CVPR} for patch optimization and their white-box evaluation procedure on KITTI 2015~\cite{Menze2015CVPR}.
We show the importance of the spatial location and analyze the flow networks' feature embeddings. Through this analysis, we can trace the adversarial patch attacks back to the classical aperture problem in optical flow. For sake of brevity, we focus on FlowNetC, since it is the most vulnerable flow network (Table~\ref{tab:additionalResults}), as well as PWC-Net and RAFT. See Supplement Section \ref{sec:evalDetails} for all implementation details.

\subsection{Spatial Location Heat Map}
\begin{figure*}
    \centering
    \includegraphics[width=0.9\textwidth]{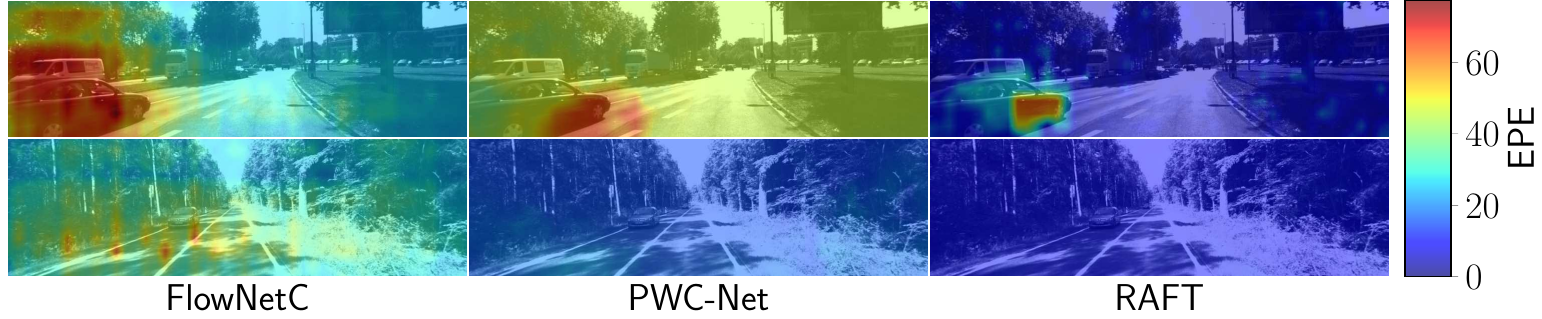}
    \caption{\textbf{Impact of spatial locations.} Effect of the spatial location of an adversarial $102\!\times\!102$ patch. Best viewed in color.}
    \label{fig:impactLocation}
\end{figure*}
We analyze the impact of the spatial location of the adversarial patch by computing the attacked End-Point-Error (EPE) for each location over a coarse grid on the image.
For visualizations of the resulting heat map, we linearly interpolate between values and clip them.
This allows us to identify three potential attacking scenarios: best case, median case, and worst case. For example, in the worst-case scenario, we paste the patch at the image location with the highest attacked EPE.
Figure~\ref{fig:impactLocation} shows that the sensitivity to patch-based attacks depends much on the image and the location of the patch.
The sensitivity of PWC-Net and RAFT can also sometimes be high.
In particular, image regions with large flow (\eg, fast-moving objects) can lead to a severe deterioration of flow estimations. 

\subsection{Correlations and Correlation Layer}
\begin{figure}
    \centering
    \begin{subfigure}[b]{0.47\textwidth}
        \centering
        \includegraphics[trim=0 0 0 0, width=0.49\textwidth]{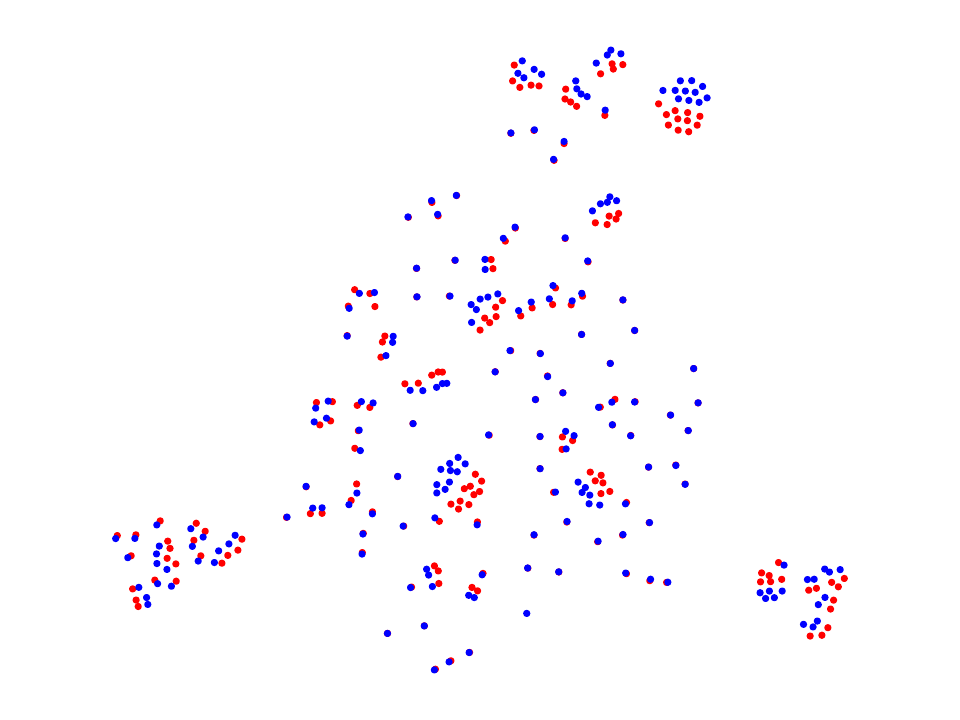}
        \includegraphics[width=0.49\textwidth]{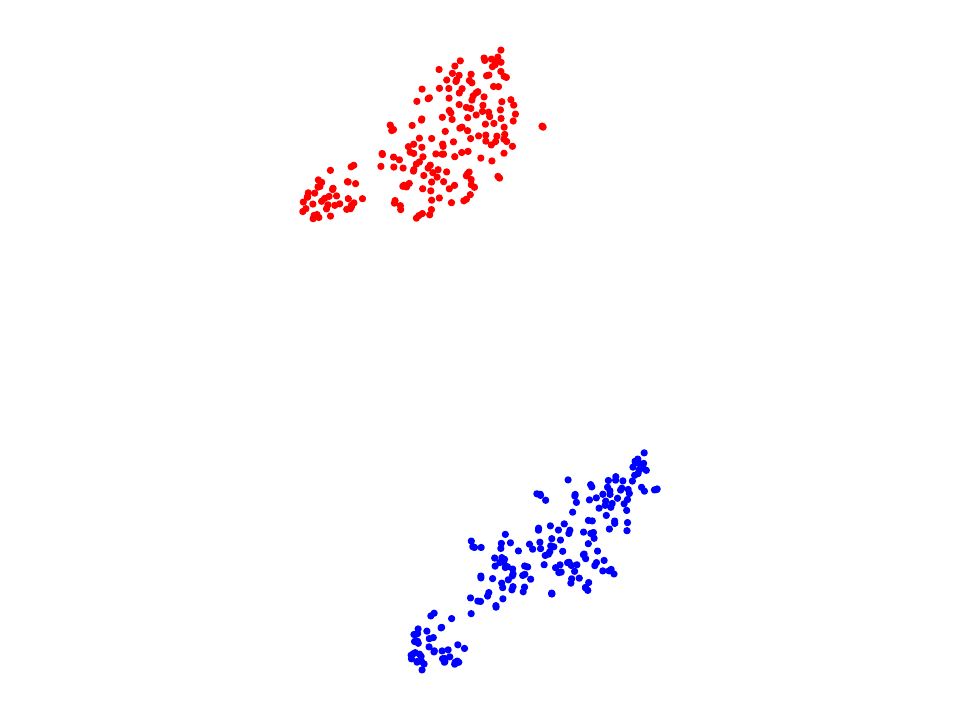}
        \caption{FlowNetC~\cite{dosovitskiy2015flownet}, MMD before: 0.246, MMD after: 3.331.}
    \end{subfigure}
    \hfill
    \begin{subfigure}[b]{0.47\textwidth}
        \centering
        \includegraphics[width=0.49\textwidth]{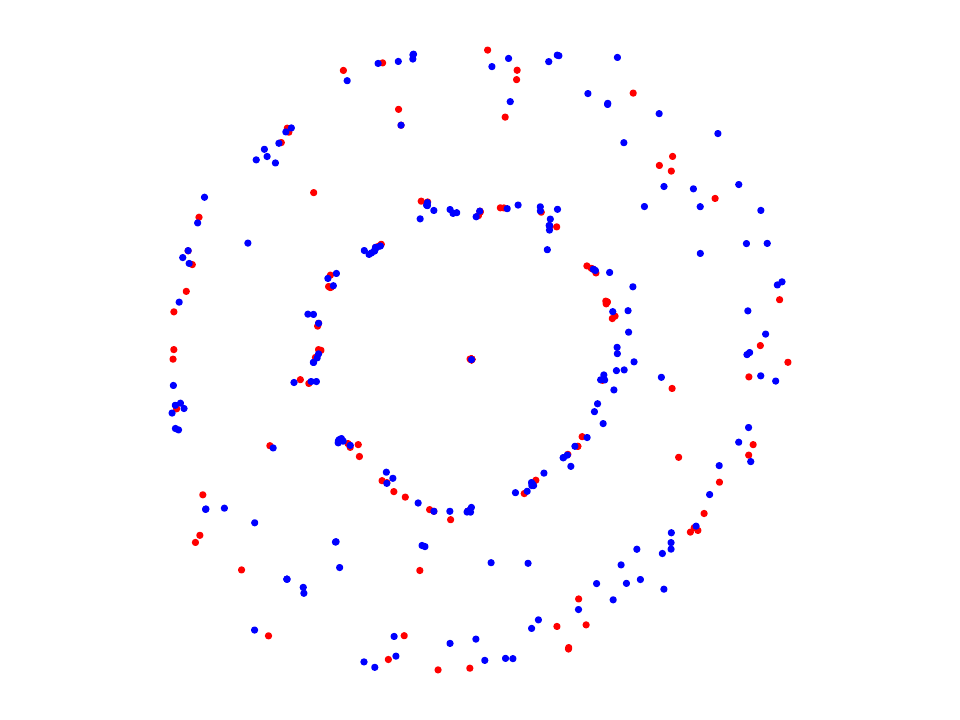}
        \includegraphics[width=0.49\textwidth]{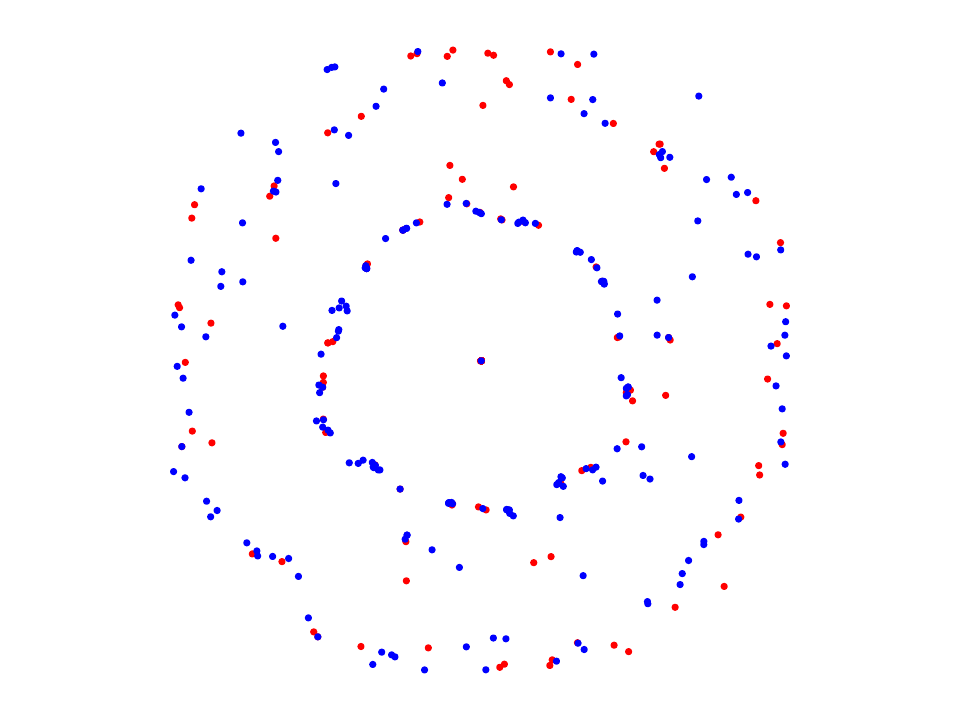}
        \caption{PWC-Net~\cite{sun2018pwc}, MMD before: 0.0, MMD after: 0.0.}
    \end{subfigure}
    \begin{subfigure}[b]{0.47\textwidth}
        \centering
        \includegraphics[width=0.49\textwidth]{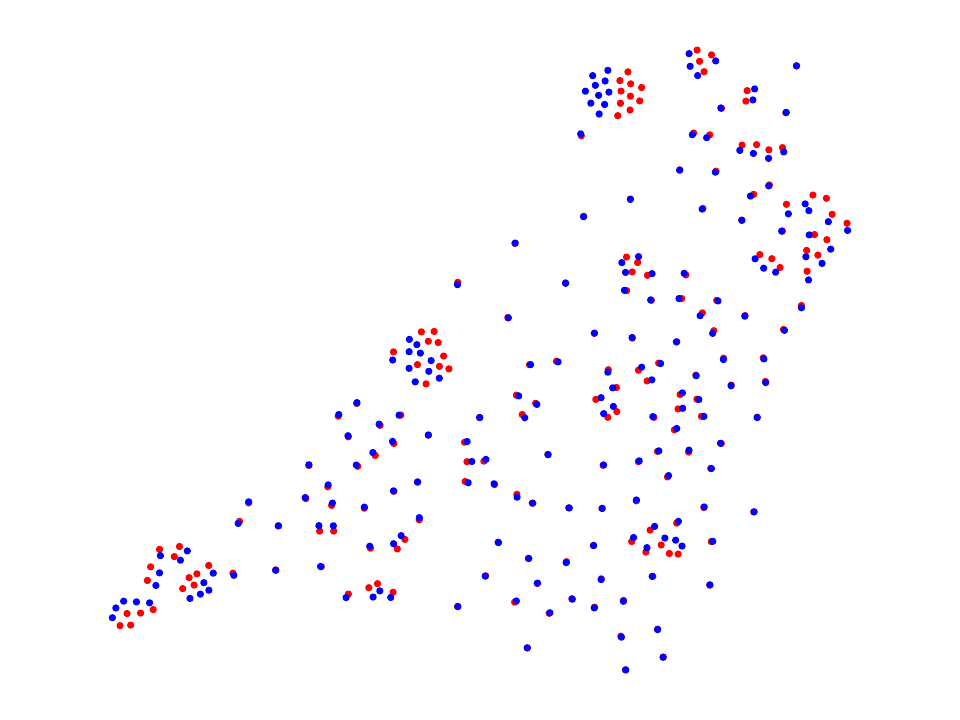}
        \includegraphics[width=0.49\textwidth]{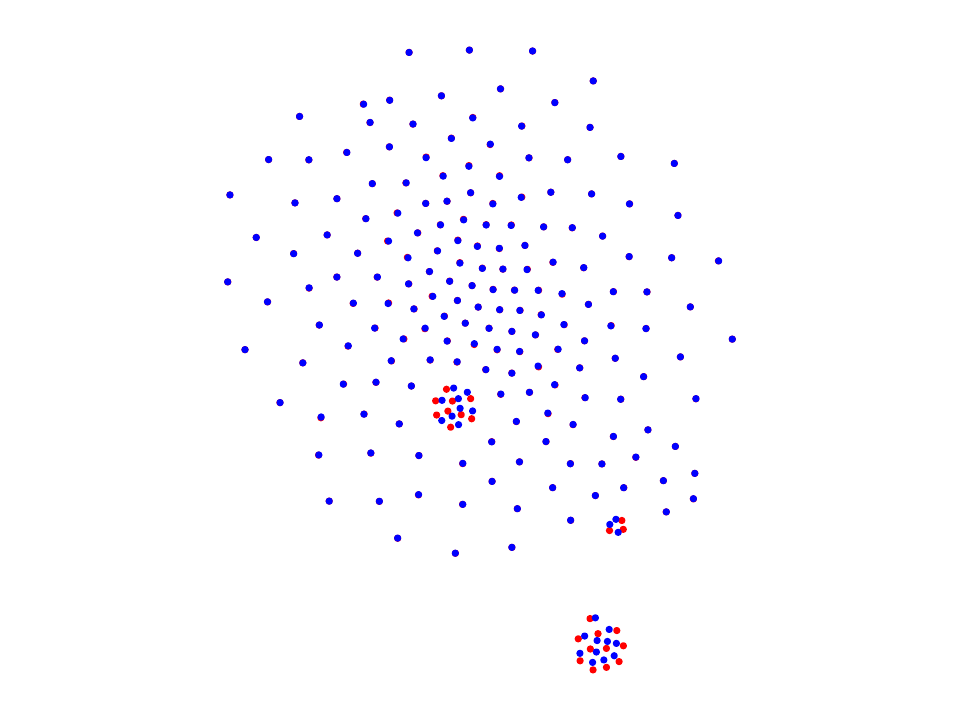}
        \caption{RAFT~\cite{teed2020raft}, MMD before: 0.158, MMD after: 0.003.}
    \end{subfigure}
    \caption{\textbf{t-SNE embeddings of features from FlowNetC, PWC-Net and RAFT.} Left: t-SNE embeddings of features before correlation layer. Right: t-SNE embeddings of features after correlation layer. We use our best found adversarial $102\!\times\!102$ patch ($2.1\%$ of the image size). Blue and red points correspond to unattacked or attacked features, respectively. We visualize the t-SNE embeddings of features of PWC-Net and RAFT before or after applying the correlation layer for the first time. Note that a larger MMD indicates that the unattacked and attacked features are more separable. Best viewed in color and with zoom.}
    \label{fig:embeddings}
\end{figure}
To analyze the features of flow networks during the attack, we visualize the unattacked and attacked features' distributions using t-SNE \cite{van2008visualizing} and compute the Maximum Mean Discrepancy (MMD) \cite{gretton2006kernel} between the two distributions.
Comparing FlowNetC's feature embeddings with and without the attack reveals a large separation of the unattacked and attacked feature distributions after the correlation layer (Figure~\ref{fig:embeddings}), while the distributions were quite close before that layer.  
This is also indicated by the rapid increase of MMD from $0.246$ to $3.331$.
On the other hand, the unattacked and attacked feature distributions of PWC-Net and RAFT are close to each other before and after applying the correlation layer, and also the MMDs stay similar. 
Hence, we hypothesize that the feature correlation of FlowNetC causes the vulnerability to patch-based attacks.

We validate this hypothesis by replacing attacked features with unattacked features to simulate what happens if an architectural component of FlowNetC would be robust \wrt patch-based attacks. We used a $102\!\times\!102$ patch with uniform noise, pasted it at a random location, and saved the feature maps. Afterward, we attacked FlowNetC with an adversarial patch of the same size at the same location and replaced the attacked feature maps with the previously saved unattacked feature maps for different architectural components. 
Table~\ref{tab:proofOfConcept} shows that a robust correlation layer (corr) could remove the effect of the attack. Trivially, a robust encoder before the correlation layer (\mbox{conv3\textlangle a,b\textrangle}) can do the same. In contrast, if the convolution that bypasses the correlation layer (conv\_redir) is made robust, the attack still remains fully effective. This shows that the feature correlation is the root cause, and also explains why FlowNetC's sibling FlowNetS is much more robust, as it has no correlation layer.
\begin{table}
    \centering
    \caption{\textbf{Replacing attacked features.} Average EPE of the attacked FlowNetC (left) and the average EPE when the respective features are replaced by those from the unattacked FlowNetC (right) on the KITTI 2015 training dataset using adversarial and uniform noise $102\!\times\!102$ patches, respectively. See Figure~\ref{fig:flowNetCEncoder} left for the encoder before the correlation layer in the original FlowNetC~\cite{dosovitskiy2015flownet}.
    }
    \resizebox{0.75\columnwidth}{!}{%
    \begin{tabular}{l|cc}
          Replace & Without & With\\
          features of & replacement & replacement \\
          \midrule
          conv3\textlangle a,b\textrangle & 25.95 & 11.31 \\
          conv\_ redir & 25.95 & 28.36 \\
          corr & 25.95 & 12.67 
    \end{tabular}
    }
    \label{tab:proofOfConcept}
\end{table}

\subsection{Relationship to the Aperture Problem}
While we have identified the correlation layer as the cause on the network side, we do not yet know what is causing it in the images.
There is good reason to suspect that the attack builds on \textit{self-similar patterns} within the adversarial patch; and indeed, they contain multiple self-similar patterns (Figure~\ref{fig:teaser}).
This suggests that patches trigger matching ambiguities that show as a large active area in the correlation output. 
Successive layers, that are supposed to interpret this output, successively spread the dominating ambiguous signals into the wider neighborhood, whereas the true correlation is outnumbered. 
This is related to the well-known aperture problem in optical flow, where repetitive patterns lead to an ambiguity in the optical flow and the receptive field (the aperture) determines the perceived motion.   

However, why are other flow networks, \eg, PWC-Net or RAFT, which also have a correlation layer, much more robust to the attack? 
We hypothesize that higher vulnerability is due to the smaller size of FlowNetC's aperture, \ie, a smaller receptive field before the correlation layer (\ie, $31\!\times\!31$).
More specifically, the larger receptive field size at the (first) correlation layer in PWC-Net and RAFT (\ie, $631\!\times\!631$ and $106\!\times\!106$) sees also areas of the image that are not affected by the attack. In addition, RAFT uses all-pairs correlation and correlation pooling, which further increases its effective receptive field size.
We hypothesize that this helps their correlation layers to keep the correlation peaked.

\section{Can We Attack Without Optimization?}\label{sec:unoptimizedAttack}
\begin{figure}
    \centering
    \includegraphics[width=0.23\textwidth]{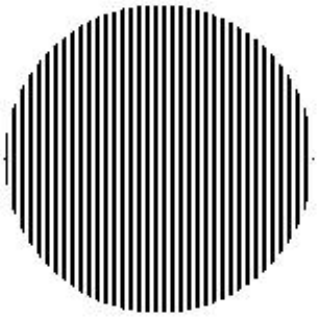}
    \caption{\textbf{Handcrafted patch.} Patch is enlarged for visualization.}
    \label{fig:unoptimizedPatch}
\end{figure}
\begin{figure*}
    \centering
    \includegraphics[width=0.94\textwidth]{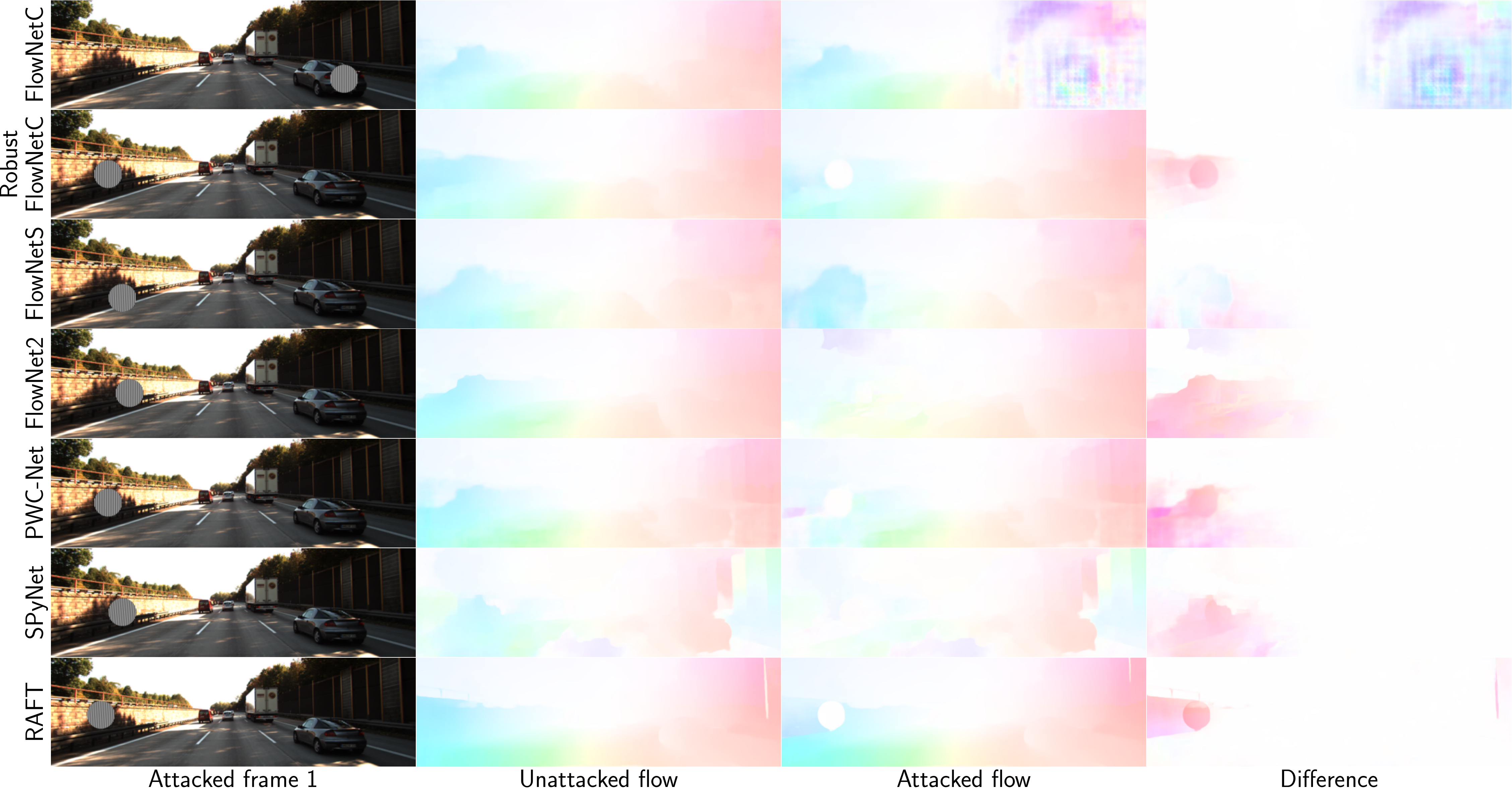}
    \caption{\textbf{Handcrafted patch attack.} Handcrafted $102\!\times\!102$ patch attack on all flow networks. We show the patch at the worst possible spatial location for each flow network. Robust flow networks predict zero flow (white color) at the patch location (third column). See Supplement Section \ref{sec:handcraftedExamples} for additional examples.}
    \label{fig:unoptimizedPatchAttack}
\end{figure*}
\begin{table*}
    \centering
    \caption{\textbf{Handcrafted patch attack.} Effect of the handcrafted patch attack (in pixel and percent of image size) on different flow networks. We show average median and worst-case attacked EPE over a coarse grid of spatial locations of the patch on the KITTI 2015 training dataset for each flow network.
    }
    \resizebox{0.85\textwidth}{!}{%
    \begin{tabular}{l|S[table-format=2.2] *{4}{|*{2}{S[table-format=2.2]}}}
          Flow & {Unattacked} & \multicolumn{2}{c|}{25x25 (0.1\%)} &  \multicolumn{2}{c|}{51x51 (0.5\%)} & \multicolumn{2}{c|}{102x102 (2.1\%)} & \multicolumn{2}{c}{153x153 (4.8\%)} \\
         Network & {EPE} & {Median} & {Worst} & {Median} & {Worst} & {Median} & {Worst} & {Median} & {Worst} \\
         \midrule
         FlowNetC \cite{dosovitskiy2015flownet} & 11.50 & 11.66 & 16.66 & 15.81 & 29.08 & 23.41 & 46.12 & 30.97 & 52.27 \\
         Robust FlowNetC & 9.95 & 9.95 & 11.14 & 9.86 & 11.74 & 9.60 & 13.08 & 9.27 & 13.64 \\
         FlowNetS \cite{dosovitskiy2015flownet} & 14.33 & 14.35 & 15.66 & 14.50 & 17.00 & 14.64 & 20.10 & 14.55 & 22.32 \\
         FlowNet2 \cite{ilg2017flownet} & 10.07 & 10.11 & 13.80 & 10.56 & 19.10 & 12.08 & 21.63 & 13.84 & 24.35 \\
         SPyNet \cite{ranjan2017optical} & 24.26 & 24.22 & 26.24 & 24.06 & 27.41 & 23.28 & 27.46 & 22.20 & 26.62 \\
         PWC-Net \cite{sun2018pwc} & 12.55 & 12.54 & 14.82 & 12.45 & 16.10 & 12.02 & 16.87 & 11.42 & 16.26 \\
         RAFT \cite{teed2020raft} & 5.86 & 5.80 & 7.08 & 5.74 & 7.44 & 5.49 & 8.69 & 5.17 & 8.96 \\
    \end{tabular}
    }
    \label{tab:unoptimizedPatchAttack}
\end{table*}
To show that self-similar patterns within patches cause the vulnerability, we handcraft a circular high-frequency black and white vertically striped patch; see Figure~\ref{fig:unoptimizedPatch}.
Note that there is no need for optimization.
We ablate the ingredients of our handcrafted patch in Supplement Section \ref{sec:ingredients}.

Table~\ref{tab:unoptimizedPatchAttack} shows that FlowNetC is also vulnerable to our handcrafted patch, providing further evidence that high correlation within the patch causes matching ambiguities in the correlation layer.
The median performance of the other flow networks, also the proposed Robust FlowNetC (Section~\ref{sec:makeFlowRobust}), is not affected, as they limit the effect of the ambiguous correlation signal to its local region or can even estimate the correct zero flow motion in this region. 
However, all flow networks are affected by the patch to some degree in the worst-case scenario, \ie, when we place the patch at the worst possible location in the image frames (Table~\ref{tab:unoptimizedPatchAttack} and Figure~\ref{fig:unoptimizedPatchAttack}). This is hard to exploit in a physical attack and is similar to other optical flow estimation errors that naturally appear locally in some difficult image frames.

%% file: 4_experiments.tex
\section{Can the Vulnerability be Controlled?}\label{sec:makeFlowRobust}
\begin{figure*}
    \centering
    \includegraphics[trim=17.7cm 1.6cm 11.8cm 0.2cm, clip=true, angle=90, width=\textwidth]{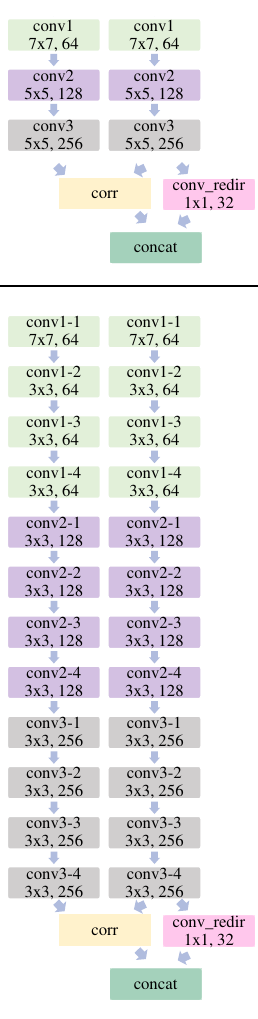}
    \caption{\textbf{Modified encoder before the correlation layer for Robust FlowNetC.} Left: original FlowNetC encoder~\cite{dosovitskiy2015flownet}. Right: our Robust FlowNetC encoder. Blocks show the name, kernel size, and number of filters. For Robust FlowNetC, the layers conv1-1, conv2-1, and conv3-1 are used for downsampling and hence have a stride of 2.}
    \label{fig:flowNetCEncoder}
\end{figure*}
Based on the previous analysis, we add corresponding architectural (and training improvements) to FlowNetC and show that this intervention also makes it robust to patch-based attacks.
The components we add to FlowNetC are already included in most modern architectures and can be regarded as the important ingredients that make an optical flow network robust to self-similar patterns as exploited by adversarial patch attacks.
Complementary, we also show that we can create a more vulnerable RAFT variant.

\subsection{Architecture}
We increase the receptive field before the correlation layer by adding (spatial resolution preserving) convolutional layers in each resolution level before the correlation layer. Moreover, we replace $5\!\times\!5$ convolutional layers in FlowNetC by $3\!\times\!3$ convolutional layers. This allows us to use deeper encoders with a larger receptive field before the correlation layer. Alternatively, we can use larger dilation rates for larger receptive fields (Supplement Section \ref{sec:dilationRate}).
We call the FlowNetC variant with kernel size $3$ and $4$ convolutional layers per resolution level \emph{Robust FlowNetC}, illustrated in Figure~\ref{fig:flowNetCEncoder} right.
For an ablation, we also created other variants of FlowNetC; see Table~\ref{tab:flowNetCVariants}.
\begin{table}
    \centering
    \caption{\textbf{Overview over FlowNetC encoder variants.} The very first layer is always a $7\!\times\!7$ convolutional layer. See Figure~\ref{fig:flowNetCEncoder} for visualizations of the original FlowNetC~\cite{dosovitskiy2015flownet} (second row) and our Robust FlowNetC encoder (last row).}
    \resizebox{0.6\columnwidth}{!}{%
    \begin{tabular}{cc|S[table-format=3]}
        Kernel & Convs per & {Receptive} \\
        size & resolution level & {field} \\
          \midrule
          3 & 1 & 19 \\
          5 & 1 & 31 \\
          3 & 2 & 47 \\
          3 & 3 & 75 \\
          5 & 2 & 87 \\
          3 & 4 & 103 \\
    \end{tabular}
    }
    \label{tab:flowNetCVariants}
\end{table}

\subsection{Training Procedure}
It has been shown that the training procedure is also an important factor for good optical flow performance~\cite{ilg2017flownet, sun2019models}.
Since we showed in the previous section that the patch-based attack is not a classical adversarial attack but simply makes the local estimation problem harder, stronger performance should also yield better robustness \wrt patch-based attacks. 
Hence, for Robust FlowNetC we used the training pipeline of RAFT, \ie, we use the AdamW optimizer~\cite{loshchilov2019decoupled}, one cycle scheduler~\cite{smith2019super}, gradient clipping, same augmentation pipeline, and also initialized the weights of the models with Kaiming initialization \cite{he2015delving}. Different from RAFT's training procedure, we used a multiscale $l_{2}$ loss, pre-train on FlyingChairs~\cite{dosovitskiy2015flownet} for $600k$ iterations with an initial learning rate of $10^{-4}$ and then trained on FlyingThings3D~\cite{mayer2016large} for $300k$ iterations with an initial learning rate of $10^{-5}$.
 
\subsection{Evaluation}
\begin{figure}
    \centering
    \begin{subfigure}[b]{0.475\textwidth}
        \centering
        \includegraphics[trim=0 0 0 0, width=\textwidth]{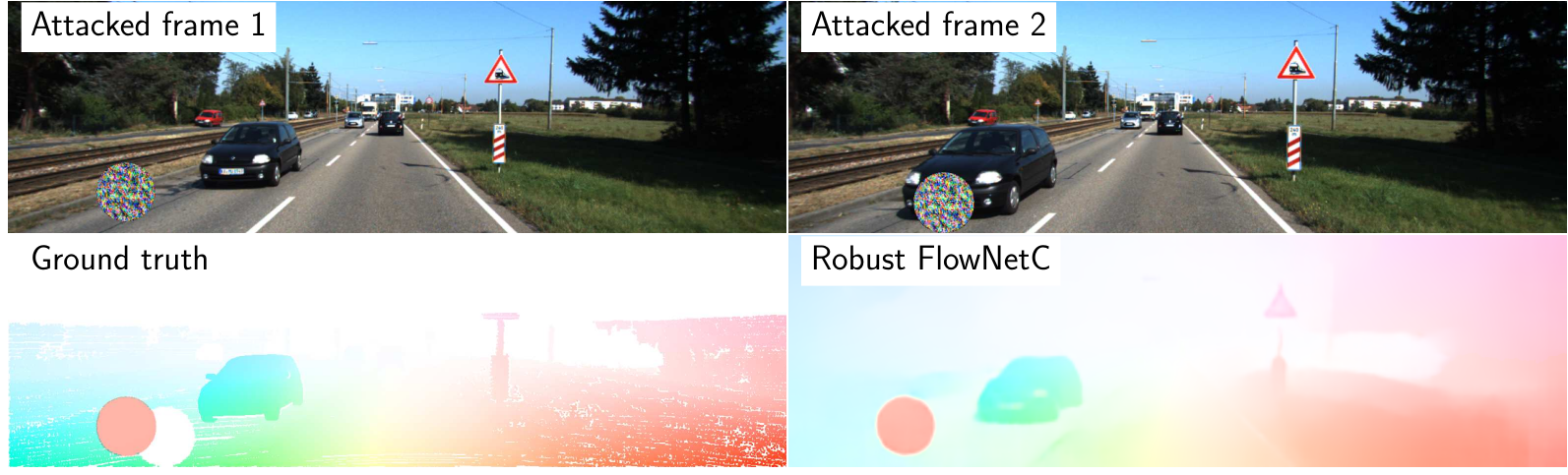}
    \end{subfigure}
    \par\bigskip
    \begin{subfigure}[b]{0.475\textwidth}
        \centering
        \includegraphics[trim=0 0 0 0, width=\textwidth]{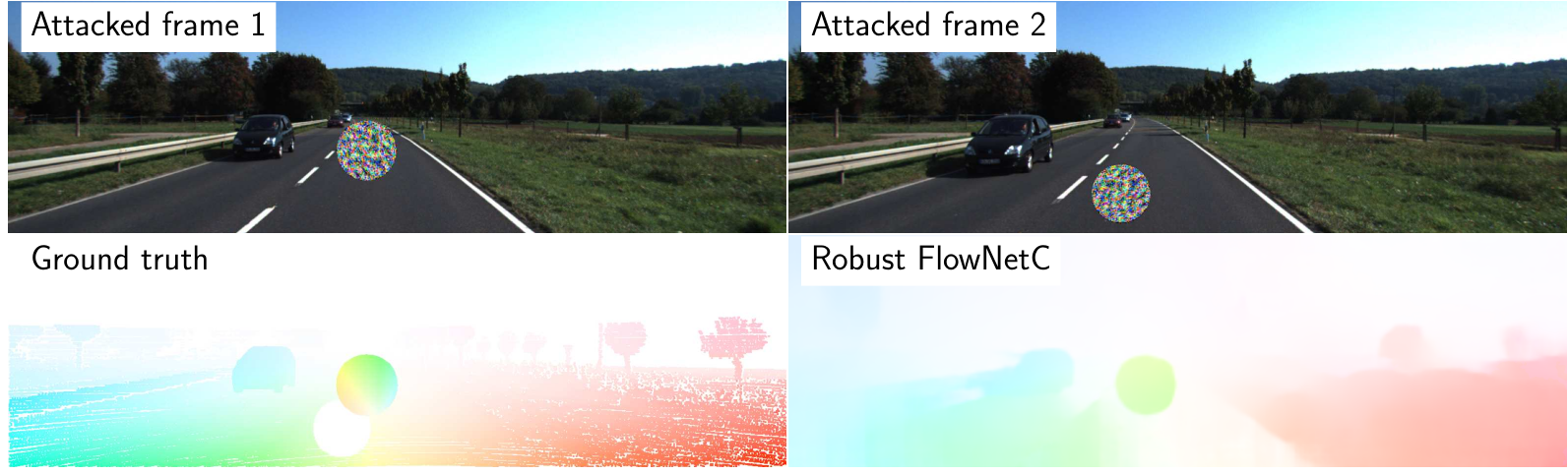}
    \end{subfigure}
    \caption{\textbf{Moving patch between image frames.} For each example block; top row shows attacked first and second image frames. Bottom row show ground truth and the predicted optical flow of Robust FlowNetC. We apply random affine transformations, \ie, translation, rotation, and scaling, to the patch between the two images frames. Note that the patch can also move in the opposite direction \wrt its neighborhood, making it even more adversarial. Robust FlowNetC correctly estimates the optical flow. Note, however, that rotations of the patch are not estimated correctly and can lead to slight estimation errors of the motion of the patch.}
    \label{fig:differentPatchPos}
\end{figure}
\begin{figure}
    \centering
    \begin{subfigure}[b]{0.475\textwidth}
        \centering
        \includegraphics[trim=0 0 0 0, width=0.49\textwidth]{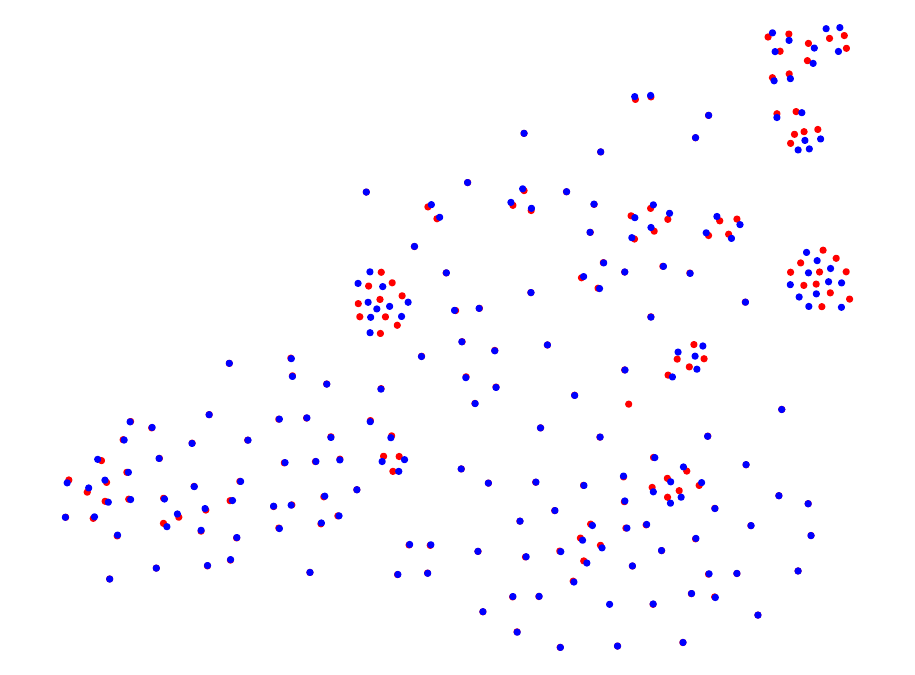}
        \includegraphics[width=0.49\textwidth]{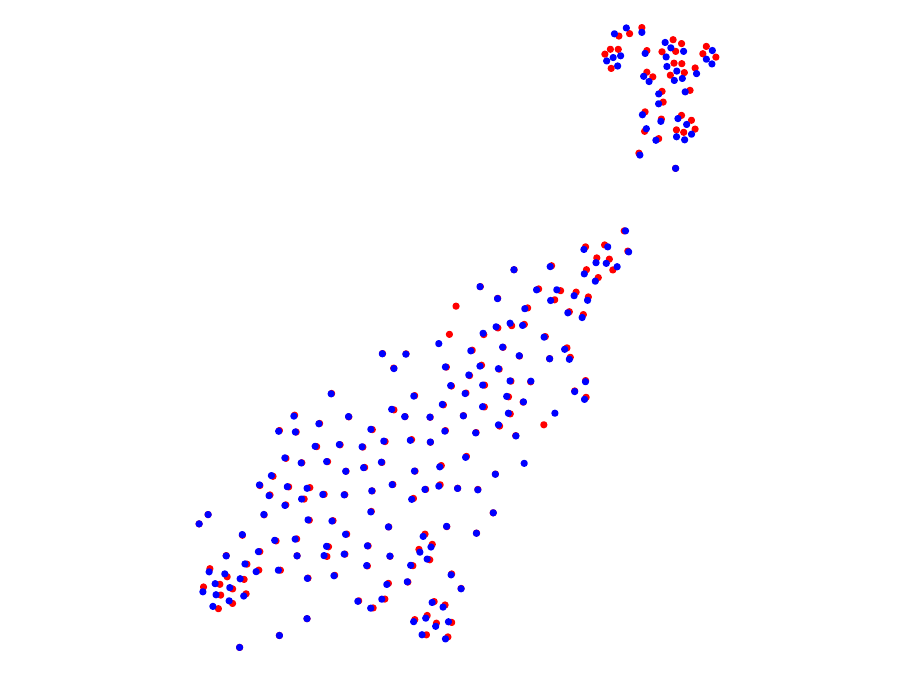}
    \end{subfigure}
    \caption{\textbf{t-SNE embeddings of features from Robust FlowNetC.} 
    Left: t-SNE embeddings of features before the correlation layer (MMD: 0.012). Right: t-SNE embeddings of features after the correlation layer (MMD: 0.007). Blue and red points correspond to unattacked or attacked features, respectively. Best viewed in color. In contrast to the original FlowNetC (Figure~\ref{fig:embeddings}a), the attacked and unattacked t-SNE embeddings stay well-aligned.}
    \label{fig:robustEmbeddings}
\end{figure}
\begin{figure}
    \centering
    \includegraphics[width=0.45\textwidth]{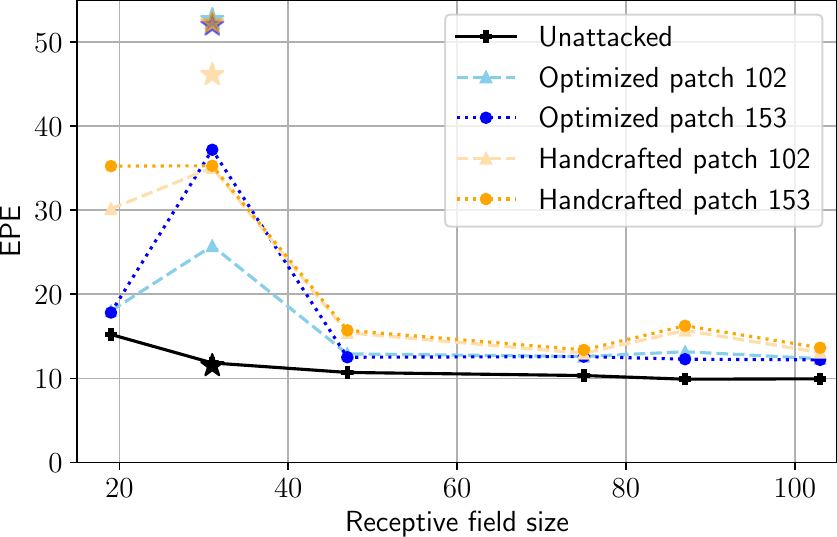}
    \caption{\textbf{Performance of FlowNetC variants with different receptive field sizes.} We show both unattacked and attacked worst-case EPE. Stars show results for the original FlowNetC. For optimized patches, we show results using the patch with the highest attacked EPE after optimization over ten runs.
    Larger receptive fields reduce the attacked worst-case EPE. 
    We report the worst-case attack \wrt location, \ie, there remains a small gap between the attacked and the unattacked result, as for all networks; see Table~\ref{tab:unoptimizedPatchAttack}.  
    The two local peaks correspond to the variants with $5\!\times5\!~$ kernel sizes; see Table~\ref{tab:flowNetCVariants}.
    }
    \label{fig:results}
\end{figure}
Figure~\ref{fig:unoptimizedPatchAttack} and Table~\ref{tab:unoptimizedPatchAttack} clearly show the effect of above changes: Robust FlowNetC is as robust to adversarial patch attacks as PWC-Net or RAFT.
The handcrafted patch attack rules out that this robustness is due to obfuscated gradients~\cite{obfuscated-gradients}.
See Supplement Section \ref{sec:robustFlowCExamples} for examples using optimized patches.
In Figure~\ref{fig:differentPatchPos}, we show a scenario where the patch is allowed to (freely) move between image frames. See Supplement Section \ref{sec:staticPatch} for results for a static patch. Figures~\ref{fig:unoptimizedPatchAttack}~and~\ref{fig:differentPatchPos} show that Robust FlowNetC correctly predicts the flow whether the patch moves or not between the image frames. The patch has only a negligible impact on the surrounding image region, even if we move the patch between image frames. 
We also tested the $l_2$ loss for patch optimization against Robust FlowNetC, and it also did not lead to any (significant) degradation in flow performance, \ie, we report worst-case EPE of $12.54$ for a $102\!\times\!102$ patch.
Figure~\ref{fig:robustEmbeddings} shows that the embeddings between the attacked and unattacked features are well-aligned -- in contrast to the original FlowNetC.
Figure~\ref{fig:results} shows that the improved robustness stems from larger receptive field sizes.

\subsection{Pushing Vulnerability}
\begin{table*}
    \centering
    \caption{\textbf{Results for our vulnerable RAFT variant.} We show average median and worst-case EPE over a coarse grid of spatial locations of our handcrafted patch on the KITTI 2015 training dataset for different RAFT variants. Even though RAFT has robust architectural ingredients, \eg, iterative refinement after the cost volume, we can substantially increase vulnerability by simple architectural changes before the cost volume.
    }
    \resizebox{0.9\textwidth}{!}{%
    \begin{tabular}{cc|S[table-format=2.2] *{4}{|*{2}{S[table-format=2.2]}}}
          FlowNetC & Without Con- & {Unattacked} & \multicolumn{2}{c|}{25x25 (0.1\%)} & \multicolumn{2}{c|}{51x51 (0.5\%)} & \multicolumn{2}{c|}{102x102 (2.1\%)} & \multicolumn{2}{c}{153x153 (4.8\%)} \\
          Encoder & {text Encoder} & {EPE} & {Median} & {Worst} & {Median} & {Worst} & {Median} & {Worst} & {Median} & {Worst} \\
         \midrule
         - & - & 5.86 & 5.80 & 7.08 & 5.74 & 7.44 & 5.49 & 8.69 & 5.17 & 8.96\\
         - & \checkmark & 6.88 & 6.78 & 9.09 & 6.65 & 9.35 & 6.43 & 10.09 & 6.10 & 11.31\\
         \checkmark & - & 5.84 & 5.85 & 7.47 & 5.84 & 9.31 & 5.78 & 10.50 & 5.92 & 11.60\\
         \checkmark & \checkmark & 6.33 & 6.38 & 13.61 & 6.43 & 16.61 & 7.03 & 19.12 & 9.37 & 20.99\\
    \end{tabular}
    }
    \label{tab:raftAttackable}
\end{table*}
\begin{figure}
    \centering
    \begin{subfigure}[b]{0.475\textwidth}
        \centering
        \includegraphics[trim=0 0 0 0, width=\textwidth]{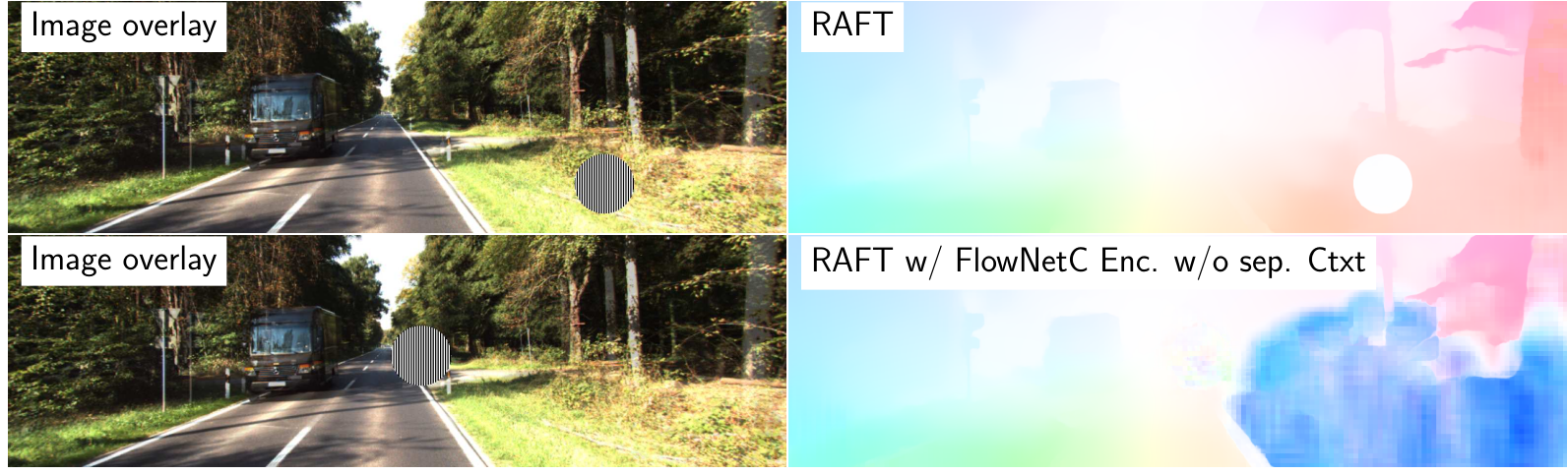}
    \end{subfigure}
    \par\bigskip
    \begin{subfigure}[b]{0.475\textwidth}
        \centering
        \includegraphics[trim=0 0 0 0, width=\textwidth]{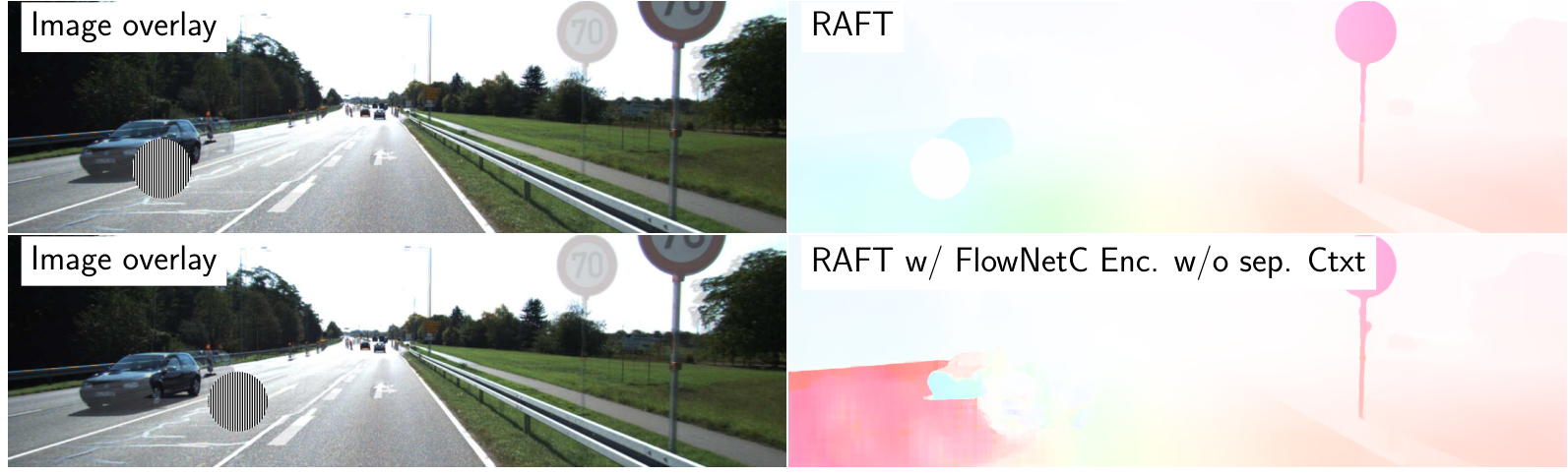}
    \end{subfigure}
    \caption{\textbf{Results for our vulnerable RAFT variant.} For each block; first column shows image overlays where we place the patch at the worst location. Second column shows the predicted optical flows of RAFT and its vulnerable variant. The vulnerable RAFT variant is vulnerable to patch-based attacks.}
    \label{fig:raftAttack}
\end{figure}
In the previous subsections, we showed that we can make FlowNetC robust by increasing its depth and, thus, its receptive field. In this section, we show the other direction by making a previously robust flow network (\ie, RAFT) vulnerable to patch-based attacks by replacing its encoder with FlowNetC's original encoder before the correlation layer (and removing the separate context encoder). Note that with these changes, the architectural part before the cost volume is the same as in FlowNetC.
We followed RAFT's training strategy~\cite{teed2020raft}.
Table~\ref{tab:raftAttackable} and Figure~\ref{fig:raftAttack} show that even with robust parts after the correlation layer, \ie, iterative refinement, there can be severe adversarial noise in the flow estimates during an attack with our handcrafted patch.

%% file: 4b_global_attacks.tex
\section{Adversarial Perturbation Attacks}\label{sec:globalAttacks}
Recently, Wong~\etal~\cite{wong2021stereopagnosia} showed that they could attack stereo networks using commonly used (global) untargeted adversarial perturbation attacks for recognition networks. 
Their approach is also effective against flow networks (Supplement Section \ref{sec:untargetedAdvAttacks}).
In the following, we propose how we can make flow networks predict any desired flow estimate by adding imperceptible adversarial perturbations, and also investigate universal perturbation attacks.
See Supplement Section \ref{sec:evalDetails} for implementation details.
Furthermore, in Supplement Section \ref{sec:adv_augmentation}, we show that we can make flow networks robust through adversarial data augmentation.

\begin{figure}
    \centering
    \includegraphics[width=0.475\textwidth]{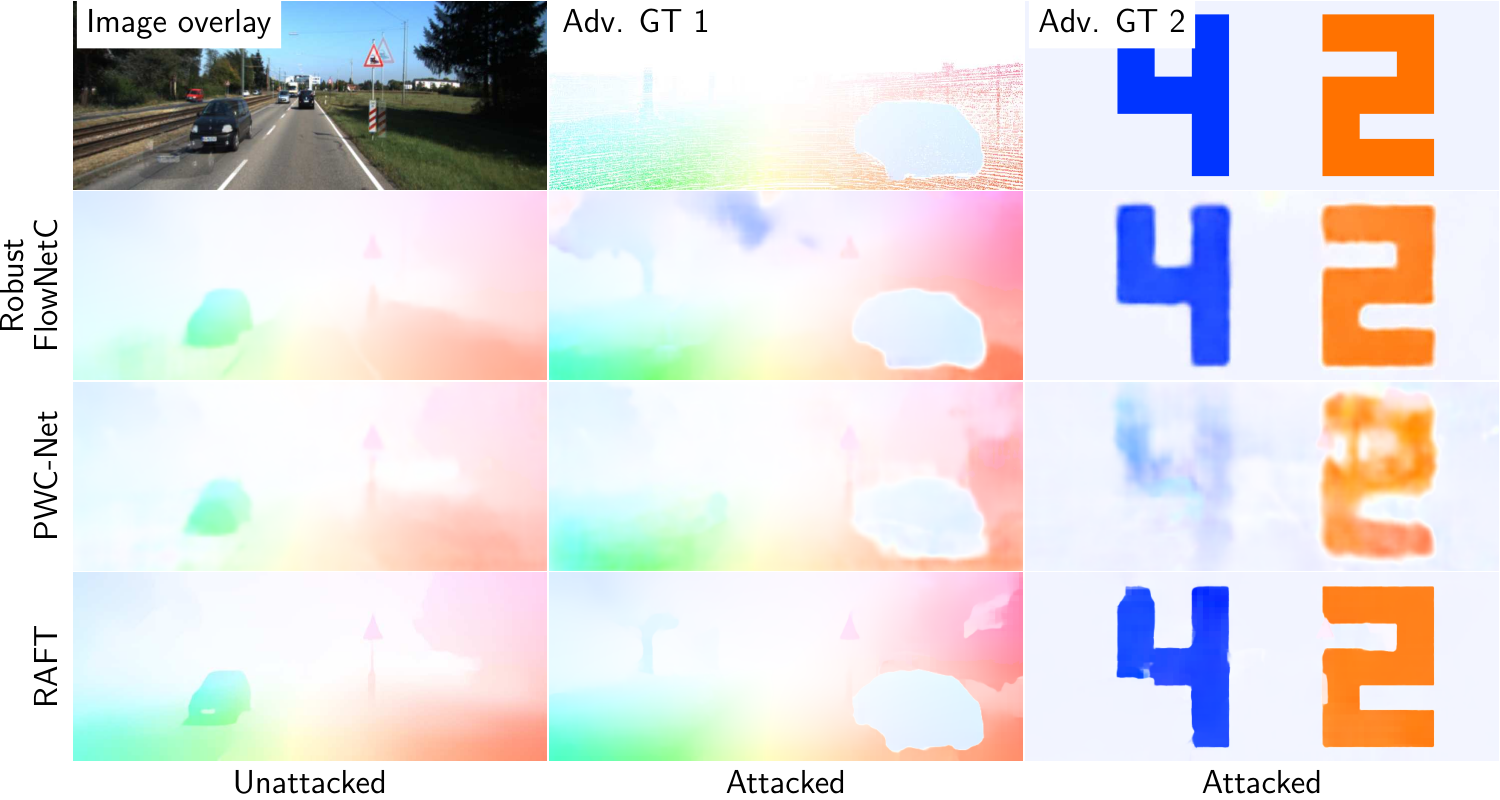}
    \caption{\textbf{Targeted adversarial attacks.}
    We can add adversarial perturbations that make flow networks predict arbitrary flows - in this case, a different flow from another scene from the KITTI 2015 training dataset or an arbitrary flow corresponding to an image with the number 42. Note that the perturbations become more effective as the number of steps of adversarial optimization increases.
    For additional examples see Supplement Section~\ref{sec:targetedAttacks}.
    } 
    \label{fig:targetedAttack}
\end{figure}

\textbf{Targeted adversarial attacks.}
While Wong~\etal showed that they can disturb stereo networks' estimations, we show that we can make flow networks predict any desired flow by adding only small additive perturbations (\eg, $L_\infty$ norm $\epsilon=0.02$).
To craft perturbations, we used the Iterative - Fast Gradient Sign Method (I-FGSM)~\cite{kurakin2016adversarial} with learning rate $\alpha=0.002$, $l_2$ loss, and minimized toward a target flow.
Figure~\ref{fig:targetedAttack} shows that we can make flow networks predict an arbitrary target flow from the same or even a completely different domain. 

\begin{figure}
    \centering
    \includegraphics[width=0.45\textwidth]{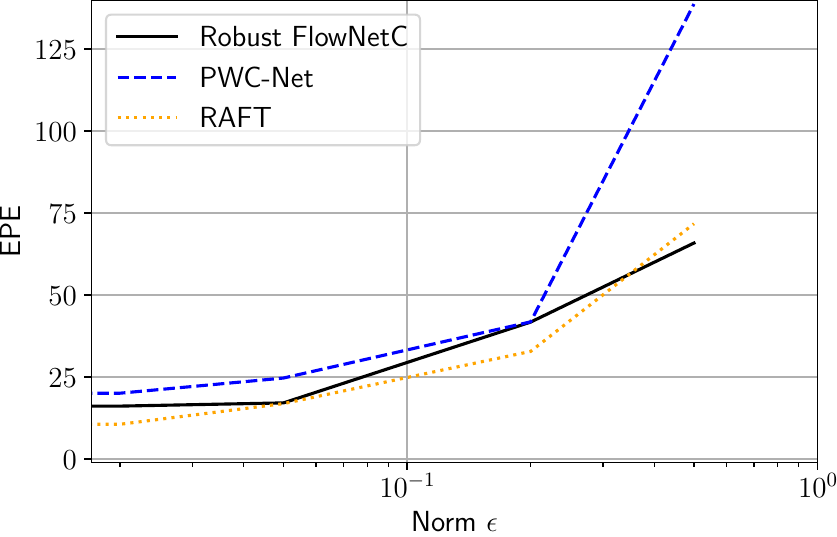}
    \caption{\textbf{Universal adversarial attacks.} We show EPE for universal adversarial attacks on KITTI 2015 training dataset for different $L_\infty$ norms (\ie, $\epsilon=\lbrace 0.0, 0.02, 0.05, 0.2, 0.5\rbrace$) and flow networks.
    See exemplary images in Supplement Section~\ref{sec:universalAttacks}.
    }
    \label{fig:universalPerturbAttack}
\end{figure}

\textbf{Universal adversarial attacks.}
We adapted the adversarial optimization of Ranjan~\etal~\cite{ranjan2019attacking} to craft universal adversarial perturbations. We used the I-FGSM attack with five steps, learning rate $\alpha=0.002$, and $l_2$ loss; all other parts remain the same as for adversarial patch optimization.
Figure~\ref{fig:universalPerturbAttack} shows that there is no severe drop in flow performance for smaller $L_\infty$ norms; only for larger $L_\infty$ norms does the flow performance drop significantly.
We find that (imperceptible) universal adversarial perturbations do not retain the severe effect of white-box adversarial attacks.

%% file: 5_conclusion.tex
\section{Discussion}
We have shown that self-similar patterns in conjunction with the correlation layer explain the vulnerability of flow networks to adversarial patch attacks. Self-similar patterns are a well-known problem for optical flow estimation and can be related to the aperture problem. In fact, we showed that a simple handcrafted self-similar patch has almost the same effect as an optimized adversarial patch. 

As we understand the cause of the problem, there is a reliable way to prevent it: 
increasing the depth, and thereby increasing the receptive field size, such that the ambiguity caused by the self-similar pattern gets resolved. Many modern networks already have a deep encoder before the correlation layer with a large enough receptive field, and, hence are robust to patch-based attacks via self-similar patterns. 
Thanks to our analysis, this is not simply a coincidence but can be explained. 

We also showed that with targeted adversarial perturbations, an attacker can produce virtually every desired flow. We also find that universal adversarial perturbations do not retain the effect of white-box adversarial attacks.
This leads to an interesting interpretation: well-designed flow networks are not vulnerable to adversarial perturbations themselves but to the superposition of image pairs and a corresponding adversarial perturbation. In practice, this means that flow networks are robust to adversarial attacks as long as attackers do not have access to the image stream.

%% file: supplement.tex
\newcommand{\beginsupplement}{%
        \setcounter{section}{0}
     }
     
\newpage 
\clearpage

\beginsupplement 

\twocolumn[{\centering{ \Huge Supplementary Material}\vspace{3ex} \medbreak \vspace{2ex}}]

\appendix

\section{Implementation Details}\label{sec:evalDetails}
We provide implementation details for different attacks and their evaluation in the following. In all our experiments, we used pre-trained models \textit{without} fine-tuning on the KITTI dataset.
We built upon the works of Ranjan~\etal~\cite{ranjan2019attacking} for adversarial patch attacks, Wong~\etal~\cite{wong2021stereopagnosia} for (global) adversarial perturbation attacks, and Teed~\etal~\cite{teed2020raft} for training of flow networks,
All code is available at \url{https://github.com/lmb-freiburg/understanding_flow_robustness}.

\paragraph{Adversarial patch attacks.}
For adversarial patch attacks, we followed the attacking and white-box evaluation procedure of Ranjan~\etal~\cite{ranjan2019attacking}.
We optimized a circular patch by optimizing \wrt Equation \ref{eq:patchOpt} using the flow networks' predictions as pseudo ground truth from the raw KITTI 2012 dataset~\cite{Geiger2012CVPR} for adversarial optimization and the annotated images as the validation set. 
We used scale augmentation within $\pm 5\%$, rotation augmentation within $\pm 10^{\circ}$ and randomly pasted the patch at different image locations, but at the \textit{same} location in both image frames.

For evaluation of patch-based experiments, we used the KITTI 2015 training set~\cite{Menze2015CVPR} and resized images to $384\times 1280$. During the evaluation, we pasted the patch also at the \textit{same} location in both image frames, if stated not otherwise. 
We always computed the unattacked and attacked End-Point-Error (EPE) and set the ground truth region occluded by the patch to zero motion.
For the computation of the spatial location heat map, we moved patches in strides of $25$ pixels in $x$- and $y$-direction to reduce the computational demands.
For the t-SNE~\cite{van2008visualizing} plots, we extracted the feature maps from the flow networks, computed the mean over the spatial dimensions to reduce the dimensionality, and computed the t-SNE embeddings on them.
For experiments with Robust FlowNetC and its variants, we optimized $>20$ or $10$ adversarial patches, respectively, across various learning rates for each patch size. We chose the three worst patches in terms of attacked EPE on the validation set, computed the spatial location heat map to get the worst-case attacked EPE and report the highest worst-case attacked EPE of the three. Other patches were not as effective as the three worst patches.
Finally, we also tested moving the patch between image frames. For this, we randomly sampled translation within $\pm 50$, full rotation (\ie $\pm 180^{\circ}$) and scale augmentation within $\pm 5\%$. 

\paragraph{Adversarial perturbation attacks.}
For adversarial perturbation attacks, we used the same procedure as Wong~\etal~\cite{wong2021stereopagnosia}, but pre-trained models were \textit{not fine-tuned} on the KITTI dataset and minimized the $l_2$ loss, as it led to more severe flow performance deterioration compared to the $l_1$ or $cossim$ losses. We used the Iterative Fast Gradient Sign Method (I-FGSM)~\cite{kurakin2016adversarial} for crafting adversarial perturbations.
For brevity, we considered only our proposed Robust FlowNetC, PWC-Net, and RAFT.
We used the KITTI 2015 training set~\cite{Menze2015CVPR} for evaluation and resized images to $256\times 640$ due to computational limitations. We used the $L_{\infty}$ norms $\epsilon=\lbrace 0.02, 0.01, 0.005, 0.002 \rbrace$ and momentum $\beta=0.47$, but with the same learning rate $\alpha=0.002$ each.

For targeted adversarial attacks, we used the same hyperparameters but minimized the $l_2$ loss. We used more steps (\ie, $100$) for the target flow depicting the number $42$.
For universal adversarial attacks, we used the same procedure as Ranjan~\etal~\cite{ranjan2019attacking} but optimized for a universal perturbation instead of a patch. We used I-FGSM with $5$ steps, and all other hyperparameters remained unchanged.

\begin{table}
    \centering
    \caption{\textbf{Adversarial patch attacks with different input data normalizations.} We show average unattacked and average worst-case attacked EPE on the KITTI 2015 training dataset.}
    \resizebox{0.82\columnwidth}{!}{%
    \begin{tabular}{l|S[table-format=3.2]|*{2}{S[table-format=3.2]}}
                  & {Un-}  & \multicolumn{2}{c}{Attacked} \\
          Network & {attacked}         & {\text{102x102}}  & {\text{153x153}}  \\
                  & {EPE}            & {\text{(2.1\%)}}  & {\text{(5.8\%)}}  \\
          \midrule
          FlowNetC as~\cite{ranjan2019attacking} & 14.52 & 94.51 & 197.00 \\ 
          FlowNetC as~\cite{ilg2017flownet} & 11.50 & 52.66 & 51.99 \\
          FlowNet2 as~\cite{ranjan2019attacking} & 11.82 & 27.59 & 43.14 \\
          FlowNet2 as~\cite{ilg2017flownet} & 10.07 & 12.40 & 13.36 \\
    \end{tabular}
    }
    \label{tab:normalization}
\end{table}
\begin{figure*}
    \centering
    \begin{subfigure}[b]{\textwidth}
        \centering
        \includegraphics[width=1.0\textwidth]{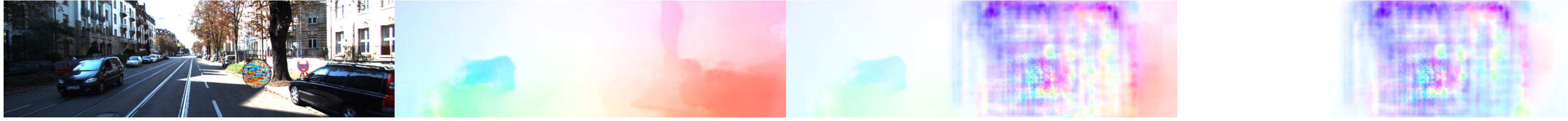}
        \caption{Normalization scheme: Ranjan~\etal~\cite{ranjan2019attacking}. Unattacked EPE: $22.71$, Attacked EPE: $60.24$.}
    \end{subfigure}
    \hfill
    \begin{subfigure}[b]{\textwidth}
        \centering
        \includegraphics[width=1.0\textwidth]{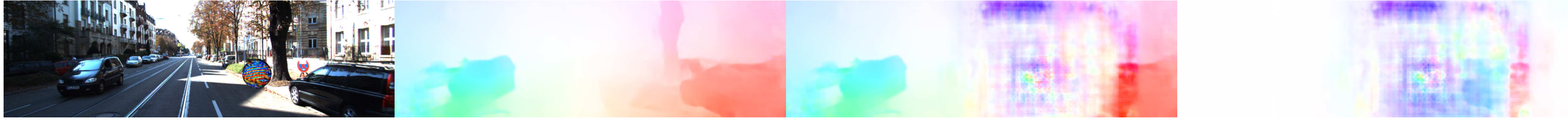}
        \caption{Normalization scheme: Ilg~\etal~\cite{ilg2017flownet}. Unattacked EPE: $16.18$, Attacked EPE: $30.03$.} 
    \end{subfigure}
    \caption{\textbf{The use of different input data normalizations leads to different results for FlowNetC.} We optimized and evaluated a $102\times 102$ patch for each input data normalization. Visualizations from left to right: the attacked first frame, the unattacked flow estimate, the attacked flow estimate, the difference between the attacked and unattacked optical flow estimates. Best viewed in color and with zoom.}
    \label{fig:normalization}
\end{figure*}
\section{Note on Input Data Normalization}
In the typical deep learning setting, we normalize the input data in the preprocessing step, as this usually leads to faster convergence~\cite{lecun2012efficient}.
Since we use input data normalization during training, we must also use the \emph{same} data normalization during inference, since the model learned based on these normalized inputs.
Using the wrong input data normalization usually has a detrimental effect on performance.

We found that Ranjan~\etal~\cite{ranjan2019attacking} normalized inputs of FlowNetC and FlowNet2 to the interval $[-1,1]$, which is different from the input data normalization FlowNetC and FlowNet2 used during their training. More specifically, Ilg~\etal~\cite{ilg2017flownet} first normalize inputs to $[0,1]$ and then subtract the mean of each RGB channel computed during the first $1000$ iterations in training.
As a result, FlowNetC's and FlowNet2's unattacked and attacked EPEs on the KITTI 2015 training dataset~\cite{Menze2015CVPR} drop significantly (Table~\ref{tab:normalization} and Figure~\ref{fig:normalization}).
However, despite this correction of the input data normalization, FlowNetC is still vulnerable, so the result of Ranjan~\etal~\cite{ranjan2019attacking} is still valid. 

\section{Additional Examples for Handcrafted Patch Attacks}\label{sec:handcraftedExamples}
\begin{figure*}
    \centering
    \begin{subfigure}[b]{0.49\textwidth}
        \centering
        \includegraphics[width=\textwidth]{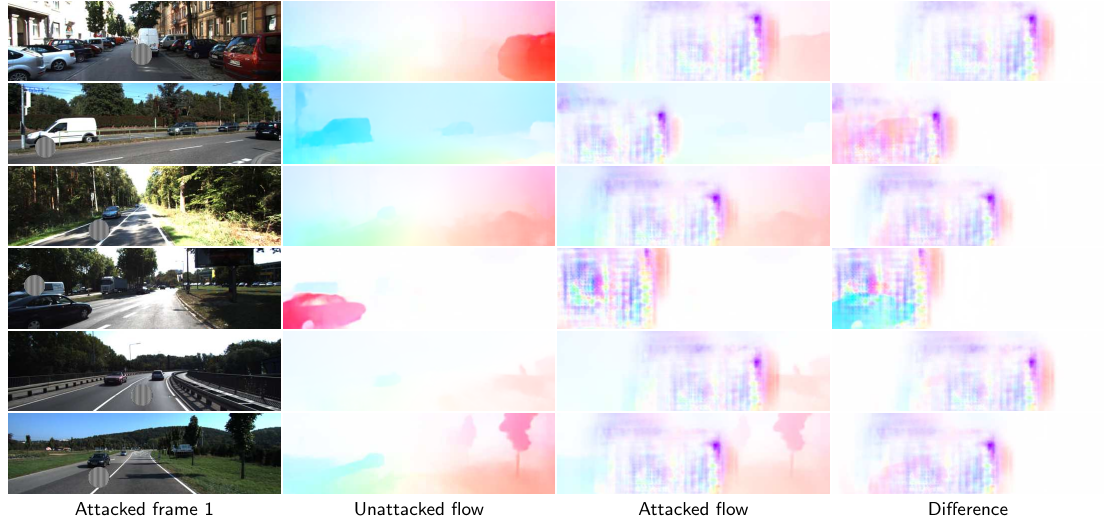}
        \caption{$102\times 102$ handcrafted patch.}
    \end{subfigure}
    \hfill
    \begin{subfigure}[b]{0.49\textwidth}
        \centering
        \includegraphics[width=\textwidth]{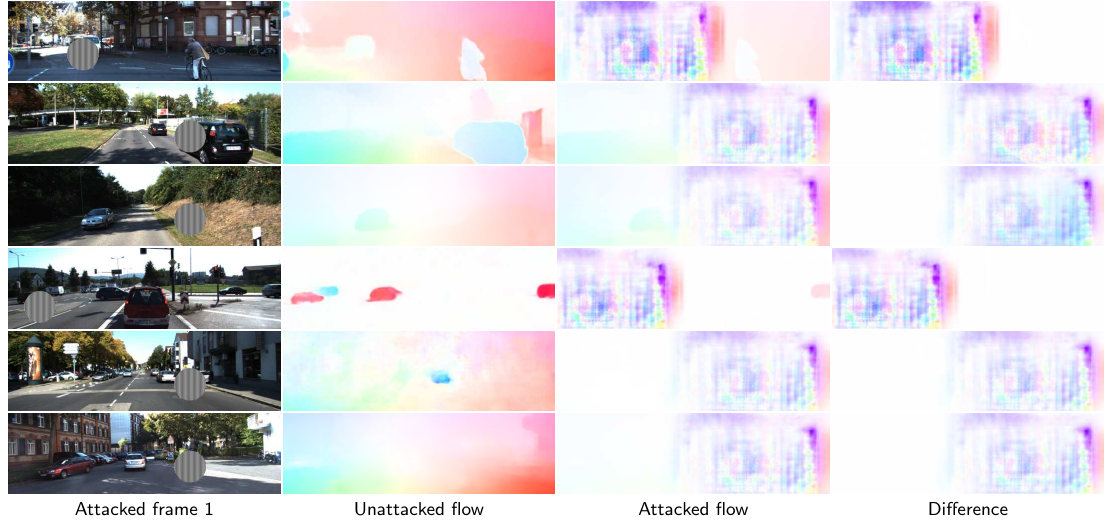}
        \caption{$153\times 153$ handcrafted patch.}
    \end{subfigure}
    \caption{\textbf{Additional examples for the handcrafted patch attack.} We show the handcrafted patch at the worst possible spatial location for FlowNetC~\cite{dosovitskiy2015flownet}. Our handcrafted patch leads to severe deteriorations of the optical flow estimates. Best viewed in color and with zoom.}
    \label{fig:handcraftedExamples}
\end{figure*}
In Figure~\ref{fig:handcraftedExamples} we show additional results for our circular high-frequency black and white vertically striped patch for FlowNetC. We only show results for FlowNetC, since it is the most vulnerable flow network and thus shows the most severe effect in the optical flow estimates. Similar to optimized patches, our handcrafted patch severely deteriorates the optical flow estimates.

\section{Ingredients for Handcrafted Patch Attacks}\label{sec:ingredients}
\begin{figure}
    \centering
    \includegraphics[width=\linewidth]{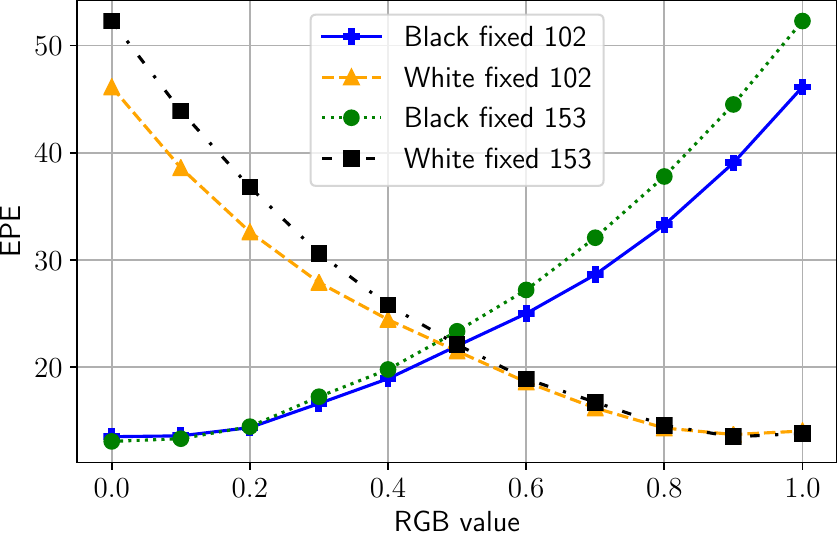}
    \caption{\textbf{Contrast between stripes of the handcrafted patch.} We fixed the black or white color of the black and white vertical striped patch fixed and moved the respective other color towards white and black, respectively. We attacked FlowNetC with these variants of our handcrafted patch. Interestingly, worst-case attacked EPE increases exponentially with contrast.}
    \label{fig:ablationContrast}
\end{figure}
\begin{figure*}
    \centering
    \begin{subfigure}[b]{0.49\textwidth}
        \centering
        \includegraphics[width=\textwidth]{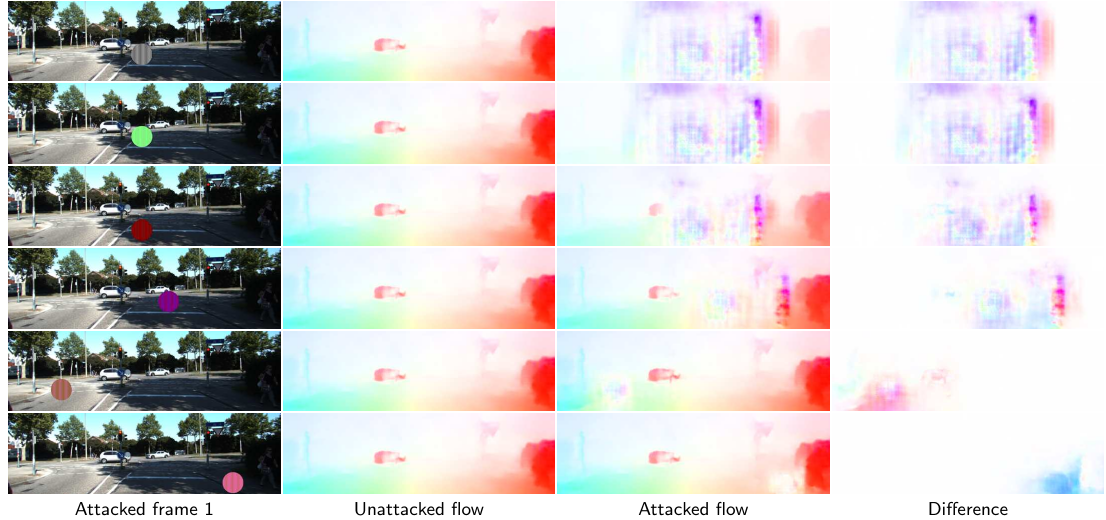}
        \caption{$102\times 102$ patches.}
    \end{subfigure}
    \hfill
    \begin{subfigure}[b]{0.49\textwidth}
        \centering
        \includegraphics[width=\textwidth]{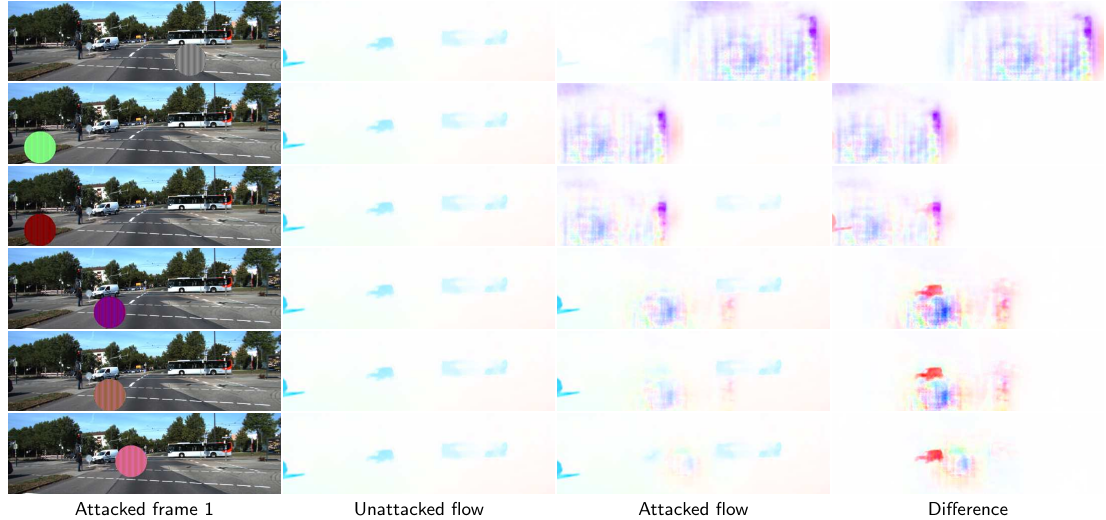}
        \caption{$153\times 153$ patches.}
    \end{subfigure}
    \caption{\textbf{Different color pairs of the handcrafted patch.} We used the same self-similar pattern with different color pairs. From top to bottom for each subfigure: black-white, green-white, red-black, red-blue, green-violet, and violet-orange. We show each handcrafted patch at the worst possible spatial location for FlowNetC~\cite{dosovitskiy2015flownet}. The handcrafted patch attack also works with different color pairs. However, the effect of the handcrafted patch may be less severe for some color pairs. Best viewed in color and with zoom.}
    \label{fig:handcraftedColor}
\end{figure*}
\begin{figure}
    \centering
    \includegraphics[width=\linewidth]{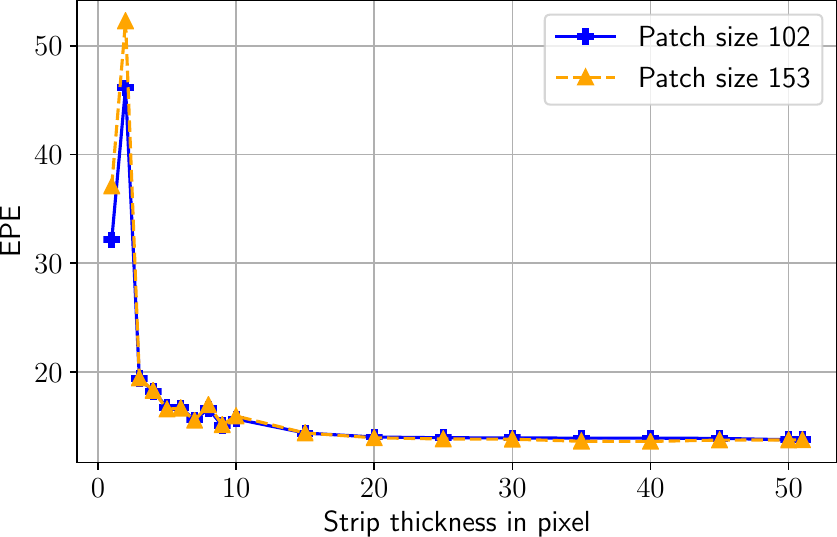}
    \caption{\textbf{Strip thickness of the handcrafted patch.} We altered the thickness of the stripes and attacked FlowNetC with these variants. The handcrafted patch requires high-frequency self-similar pattern to remain effective.}\label{fig:ablationStripThickness}
\end{figure}
\begin{figure}
    \centering
    \includegraphics[width=\linewidth]{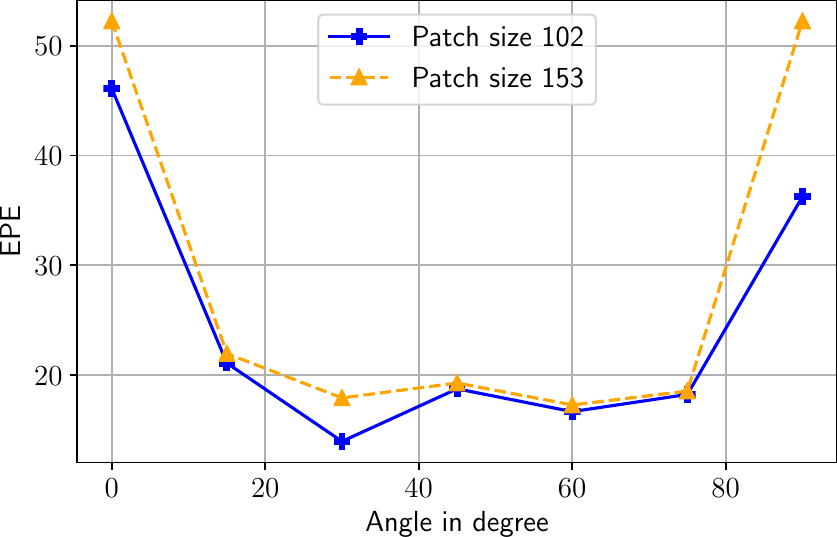}
    \caption{\textbf{Rotational orientation of the handcrafted patch.} We rotated our handcrafted patch and attacked FlowNetC. The rotational angle of $0^{\circ}$ corresponds to vertical stripes and the rotational angle $90^{\circ}$ corresponds to horizontal stripes. We observe a U-shaped form of worst-case attacked EPE: the stripes oriented in the axial directions cause more deterioration of flow performance.}\label{fig:ablationRotation}
\end{figure}
We conducted several ablations to identify the main ingredients for a successful handcrafted patch attacks besides its self-similar pattern. We chose FlowNetC as the flow network for the ablations because it is the most vulnerable \wrt patch-based attacks.
To study the influence of the contrast between the stripes, we fixed the black or white color of our handcrafted patch and changed the respective other color, thereby changing the contrast between the stripes. Figure~\ref{fig:ablationContrast} shows that higher contrasts between the stripes cause more severe deteriorations of optical flow performance. Interestingly, we observe an exponential increase in worst-case attacked EPE with the increase in the contrast between the stripes.
The handcrafted self-similar pattern also works when we use different color pairs (Figure~\ref{fig:handcraftedColor}). However, the effect of the handcrafted patch may be less severe for different color pairs. Note that regions with zero flow are more vulnerable \wrt patch-based attacks.
Furthermore, our handcrafted patch requires high-frequency self-similar patterns to remain effective (Figure~\ref{fig:ablationStripThickness}). The larger the strip thickness (\ie, the lower the frequency), the smaller the effect of the handcrafted patch attack.
Finally, the handcrafted patch attack is more effective when self-similar patterns are oriented in axial directions (Figure~\ref{fig:ablationRotation}).

\section{Increasing the Receptive Field Size by Increasing the Dilation Rate}\label{sec:dilationRate}
\begin{figure}
    \centering
    \includegraphics[width=\linewidth]{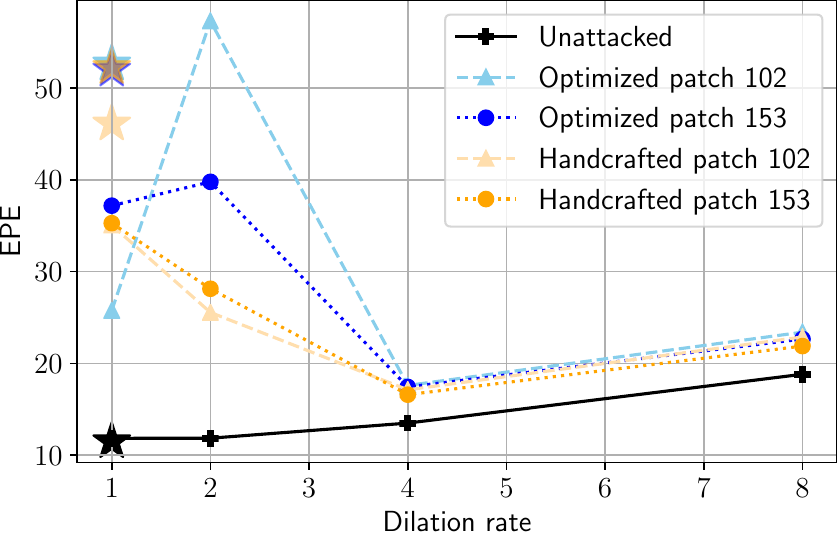}
    \caption{\textbf{Performance of FlowNetC variants with various dilation rates.} We show both unattacked and worst-case attacked EPE. Stars show results for the original FlowNetC. For optimized patches, we show results using the patch with the highest worst-case attacked EPE after optimization over ten runs. Increasing the dilation rate also improves the robustness against patch-based attacks, but overall flow performance deteriorates.}
    \label{fig:dilationExperiment}
\end{figure}
In the main paper, we showed that increasing the receptive field by increasing the network depth helps improve robustness.
Alternatively, we also tried to increase the receptive field by increasing the dilation rate of the convolutional layers of FlowNetC's encoder. We used dilation rates $\lbrace 1, 2, 4, 8\rbrace$, where a dilation rate of 1 corresponds to the original FlowNetC.
Figure~\ref{fig:dilationExperiment} shows that increasing the dilation rates also makes FlowNetC more robust \wrt patch-based attacks. The gap between unattacked and worst-case attacked EPE can be mainly attributed to occlusions causing optical flow performance to deteriorate.
However, the flow performance for unattacked image pairs deteriorates significantly at larger dilation rates. More explicitly, the FlowNetC variant with a dilation rate of 8 has unattacked EPE $18.81$ and worst-case attacked EPEs $23.4$ and $22.66$ for optimized adversarial patches with patch sizes $102\times 102$ and $153\times 153$, respectively. Note, however, that a uniform noise patch also has EPEs of $23.32$ or $22.3$ for patch sizes $102\times 102$ and $153\times 153$, respectively.
Therefore, increasing the receptive field by adding more depth is preferable to make flow networks robust \wrt patch-based attacks.

\section{Additional Examples for Robust FlowNetC}\label{sec:robustFlowCExamples}
\begin{figure*}
    \centering
    \begin{subfigure}[b]{0.49\textwidth}
        \centering
        \includegraphics[width=\textwidth]{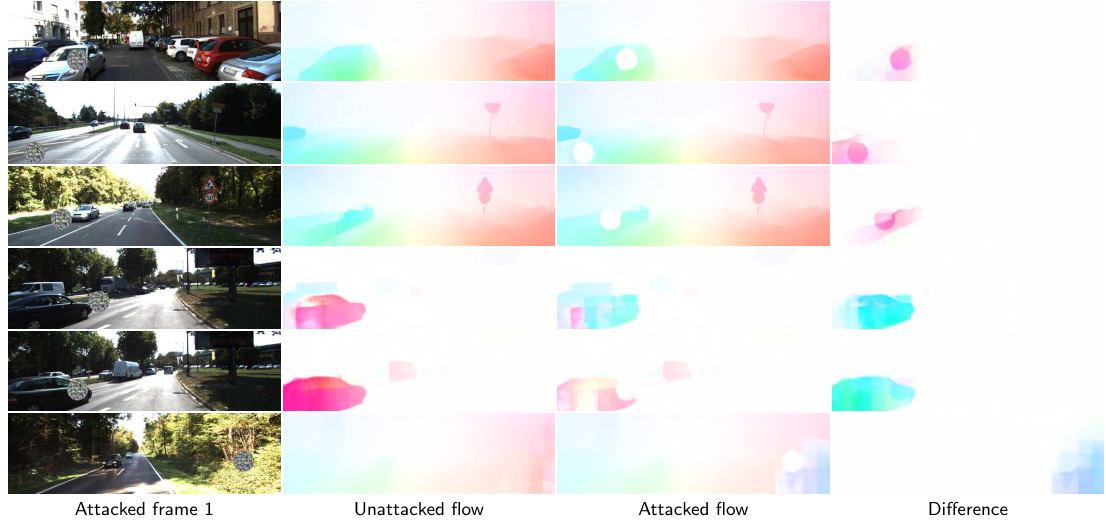}
        \caption{$102\times 102$ optimized patch.}
    \end{subfigure}
    \hfill
    \begin{subfigure}[b]{0.49\textwidth}
        \centering
        \includegraphics[width=\textwidth]{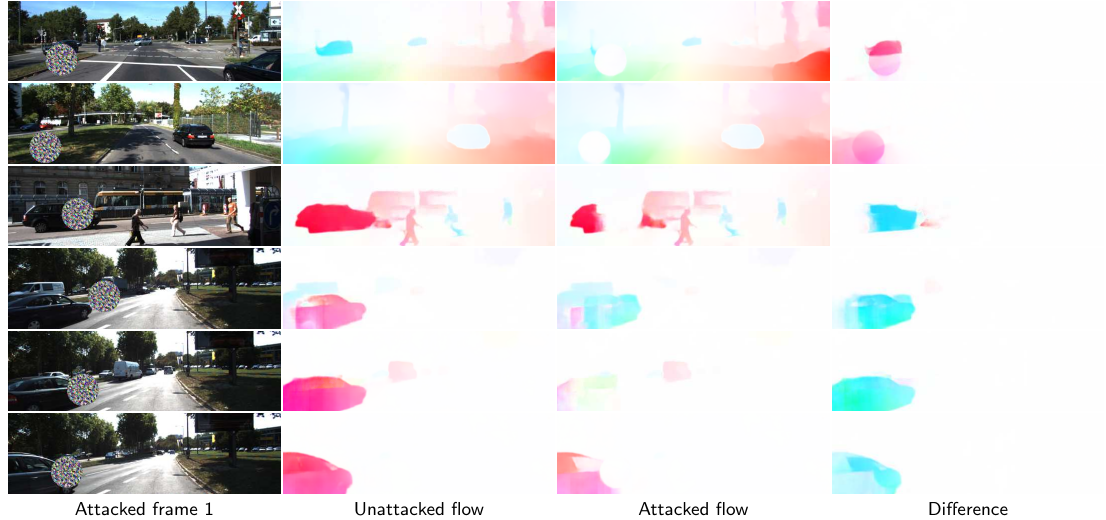}
        \caption{$153\times 153$ optimized patch.}
    \end{subfigure}
    \caption{\textbf{Additional examples for Robust FlowNetC.} We show the best found optimized patch at the worst possible spatial location for Robust FlowNetC. The bottom three rows of each subfigure show the worst examples based on the greatest absolute degradation of worst-case attacked EPE in descending order. Best viewed in color and with zoom.}
    \label{fig:robustFlowNetCExamples}
\end{figure*}
Figure~\ref{fig:robustFlowNetCExamples} shows that Robust FlowNetC is also robust against optimized patches.
However, as with other flow networks, we would like to stress that some particular hard image frames can cause severe deterioration of flow performance.

\section{Realistic Motion of Patches}\label{sec:staticPatch}
We also tried to use realistic motion of patches by considering them as part of the static scene, as described by Ranjan~\etal~\cite{ranjan2019attacking}. We found that it has a negligible effect \wrt the worst-case attacked EPE for Robust FlowNetC, \ie, $12.16$ and $12.11$ to $13.60$ and $14.57$ for $102\times 102$ or $153\times 153$ patches, respectively. 
We found the reason for higher worst-case attacked EPE is due to the placement of patches at boundary regions of the first image frame so that they disappeared in the second image frame.

\section{Untargeted Adversarial Attacks}\label{sec:untargetedAdvAttacks}
\begin{figure*}
    \centering
    \includegraphics[width=0.47\textwidth]{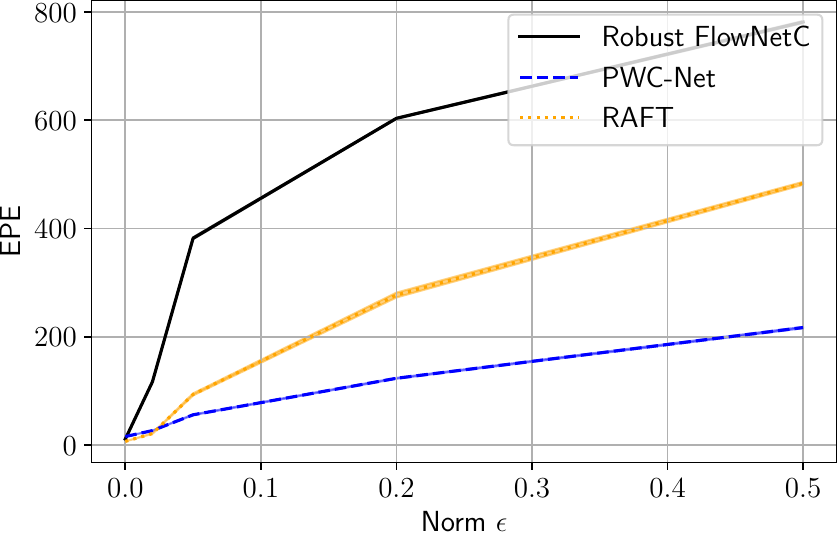}
    \caption{\textbf{Untargeted adversarial attacks.} We attacked flow networks with I-FGSM on the KITTI 2015 training dataset over various $L_\infty$ norms $\epsilon=\lbrace 0.002, 0.005, 0.01, 0.02\rbrace$. I-FGSM significantly deteriorates optical flow performance for all flow networks. 
    }
    \label{fig:untargetedAttacks}
\end{figure*}
\begin{figure*}
    \centering
    \begin{subfigure}[b]{0.49\textwidth}
        \centering
        \includegraphics[width=\textwidth]{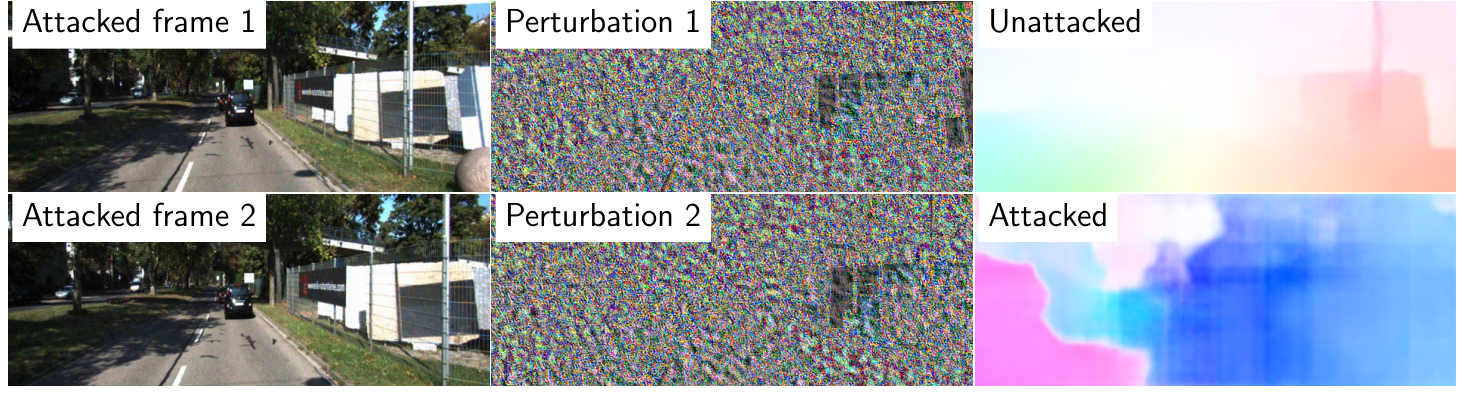}
        \caption{Robust FlowNetC.}
    \end{subfigure}
    \begin{subfigure}[b]{0.49\textwidth}
        \centering
        \includegraphics[width=\textwidth]{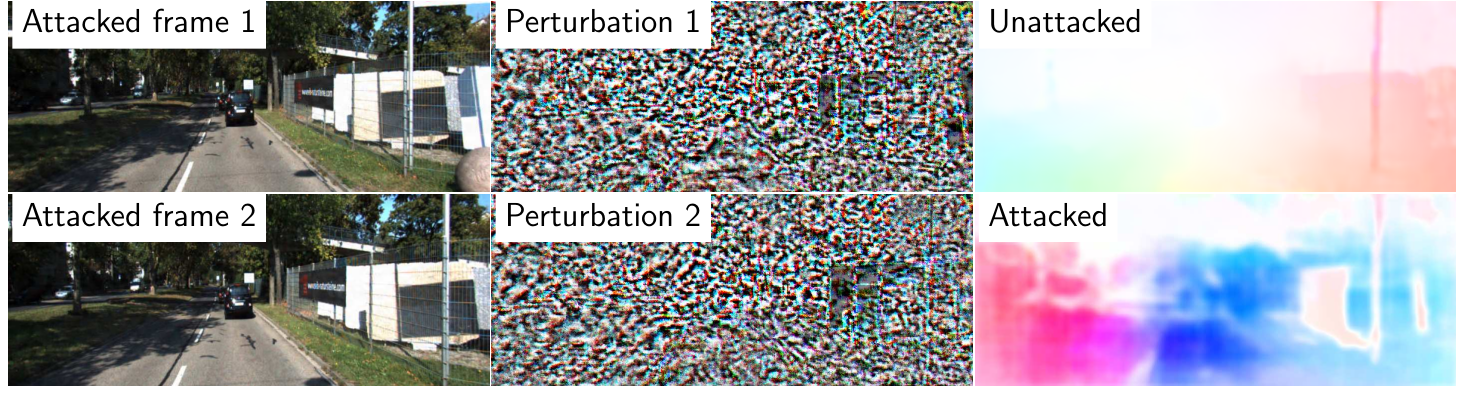}
        \caption{PWC-Net~\cite{sun2018pwc}.}
    \end{subfigure}
    \hfill
    \begin{subfigure}[b]{0.49\textwidth}
        \centering
        \includegraphics[width=\textwidth]{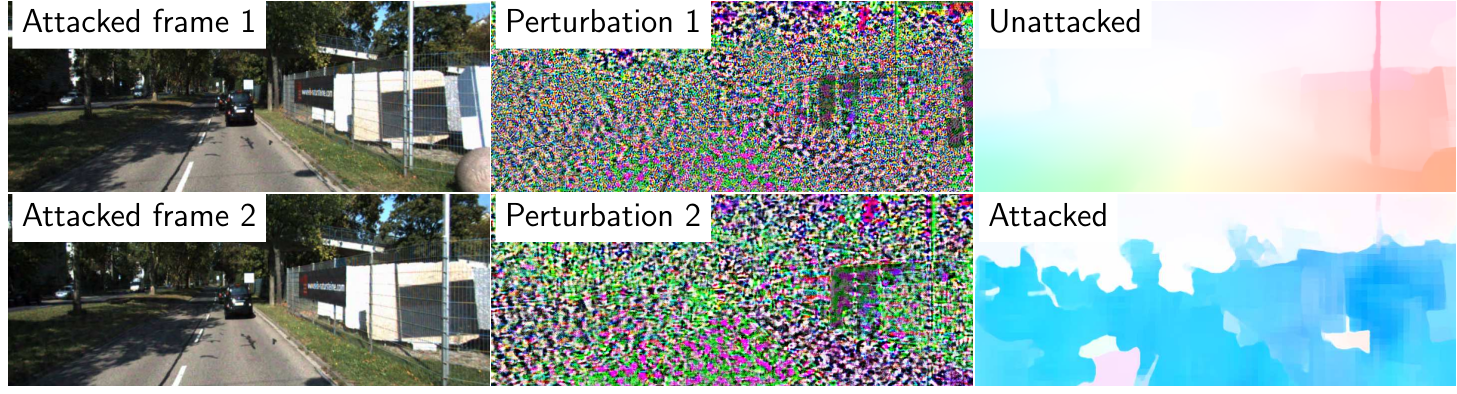}
        \caption{RAFT~\cite{teed2020raft}.}
    \end{subfigure}
    \caption{\textbf{Examples for untargeted adversarial attacks for different flow networks.} In each subfigure, we show in the first row the attacked first frame with perturbation, the perturbation of the first frame, and the unattacked flow estimate, and in the second row, the attacked second frame with perturbation, the perturbation of the second frame, and the attacked flow estimate.
    The crafted (imperceptible) adversarial perturbations completely distort the optical flow estimates. Best viewed in color and with zoom.}
    \label{fig:untargetAttackExample}
\end{figure*}
Wong~\etal~\cite{wong2021stereopagnosia} showed that they could attack disparity estimation networks by adding an adversarial perturbation individually to each pixel.
Figures~\ref{fig:untargetedAttacks} and \ref{fig:untargetAttackExample} show that, as expected, the same is true for optical flow networks. 

\section{Additional Examples for Targeted Adversarial Attacks}\label{sec:targetedAttacks}
\begin{figure*}
    \centering
    \begin{subfigure}[b]{\textwidth}
        \centering
        \includegraphics[width=0.49\textwidth]{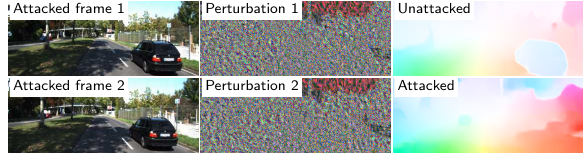}
        \includegraphics[width=0.49\textwidth]{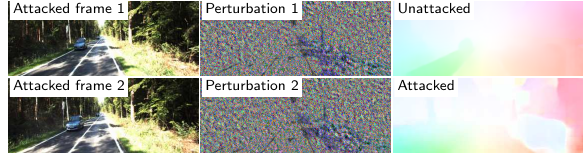}
        \includegraphics[width=0.49\textwidth]{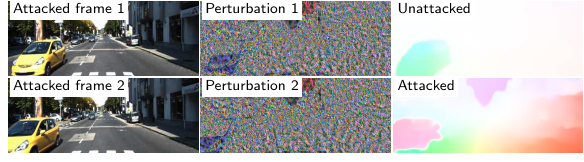}
        \includegraphics[width=0.49\textwidth]{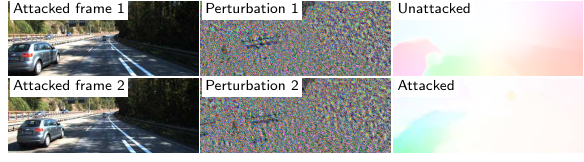}
        \caption{Robust FlowNetC.}
    \end{subfigure}
    \begin{subfigure}[b]{\textwidth}
        \centering
        \includegraphics[width=0.49\textwidth]{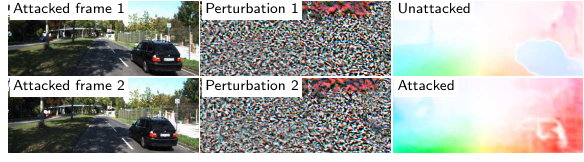}
        \includegraphics[width=0.49\textwidth]{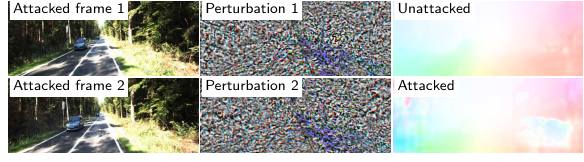}
        \includegraphics[width=0.49\textwidth]{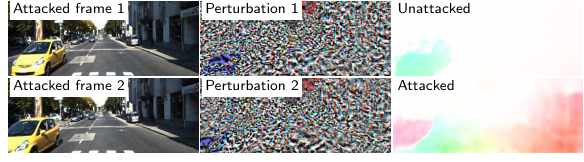}
        \includegraphics[width=0.49\textwidth]{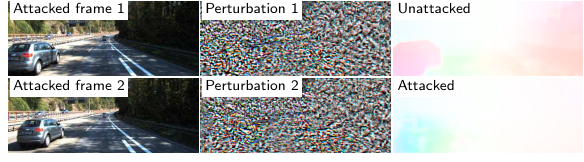}
        \caption{PWC-Net~\cite{sun2018pwc}.}
    \end{subfigure}
    \hfill
    \begin{subfigure}[b]{\textwidth}
        \centering
        \includegraphics[width=0.49\textwidth]{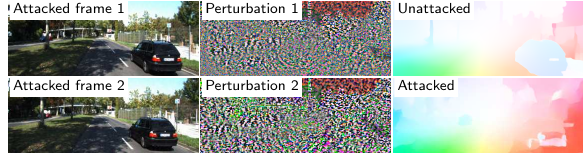}
        \includegraphics[width=0.49\textwidth]{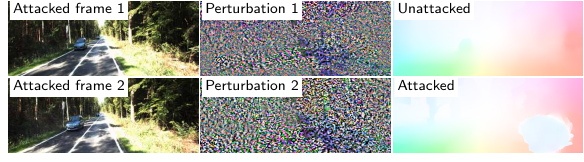}
        \includegraphics[width=0.49\textwidth]{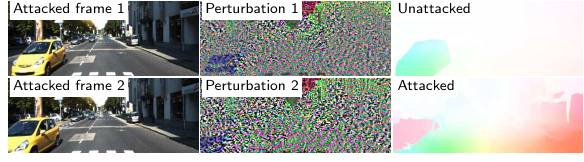}
        \includegraphics[width=0.49\textwidth]{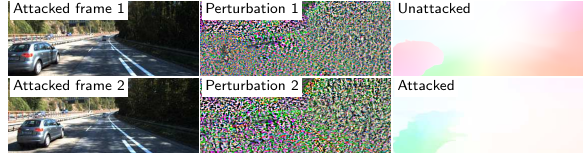}
        \caption{RAFT~\cite{teed2020raft}.}
    \end{subfigure}
    \caption{\textbf{Additional examples for targeted adversarial attacks.} 
    Targeted adversarial attacks with I-FGSM with $L_\infty$ norm $\epsilon =0.02$.
    Each row in each subfigure, the ground truth of the left or right block, is the adversarial target flow of the right or left block.
    For each block in each subfigure, we show both attacked image frames with perturbations in the first column, the perturbations for both image frames in the second column, and the unattacked and attacked flow estimate in the third column.
    Note that the flow estimates are closer to the target flows than the actual flows.
    }
    \label{fig:targetedExamples}
\end{figure*}
\begin{figure*}
    \centering
    \begin{subfigure}[b]{0.48\textwidth}
        \centering
        \includegraphics[width=\textwidth]{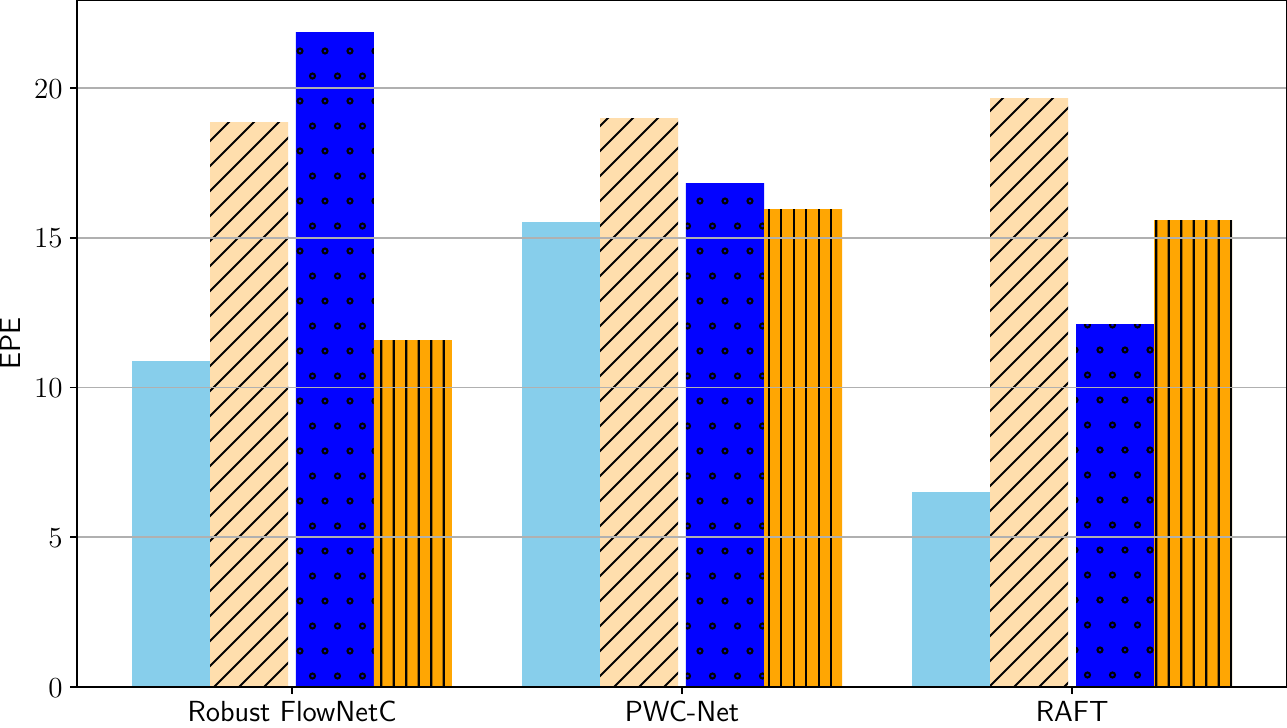}
        \caption{$\epsilon=0.002$}
    \end{subfigure}
    \begin{subfigure}[b]{0.48\textwidth}
        \centering
        \includegraphics[width=\textwidth]{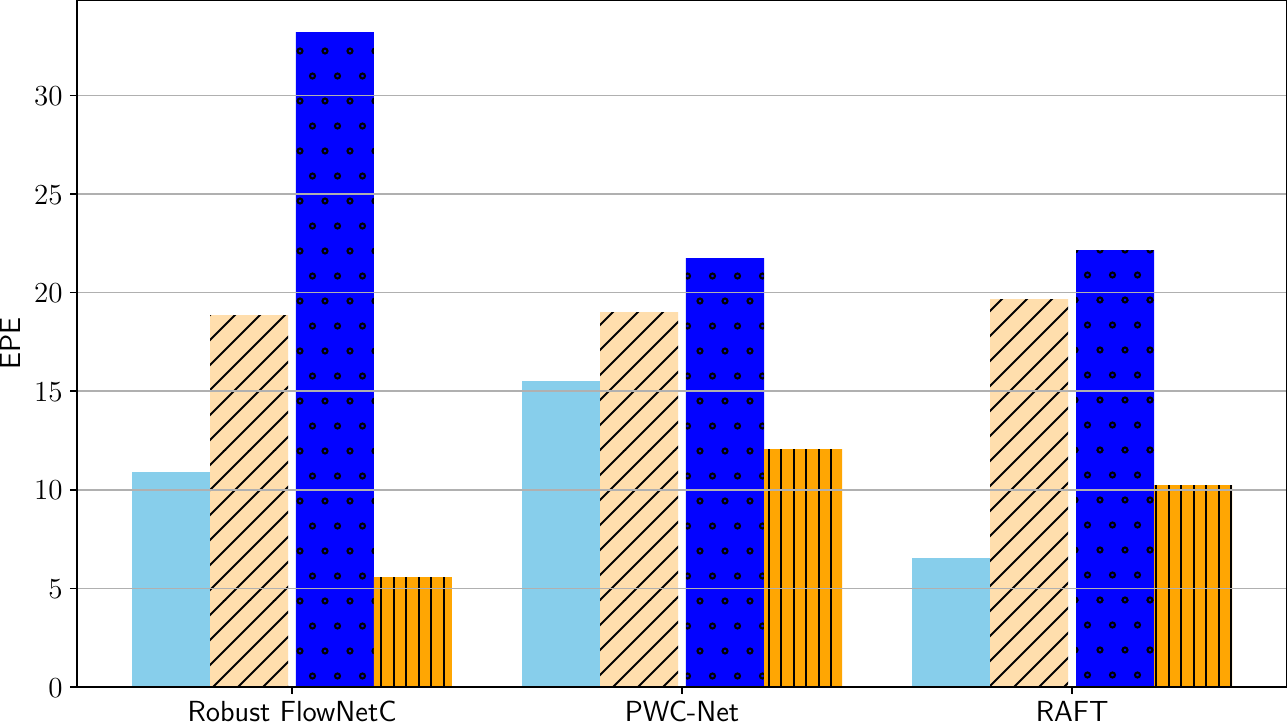}
        \caption{$\epsilon=0.005$}
    \end{subfigure}
    \begin{subfigure}[b]{0.48\textwidth}
        \centering
        \includegraphics[width=\textwidth]{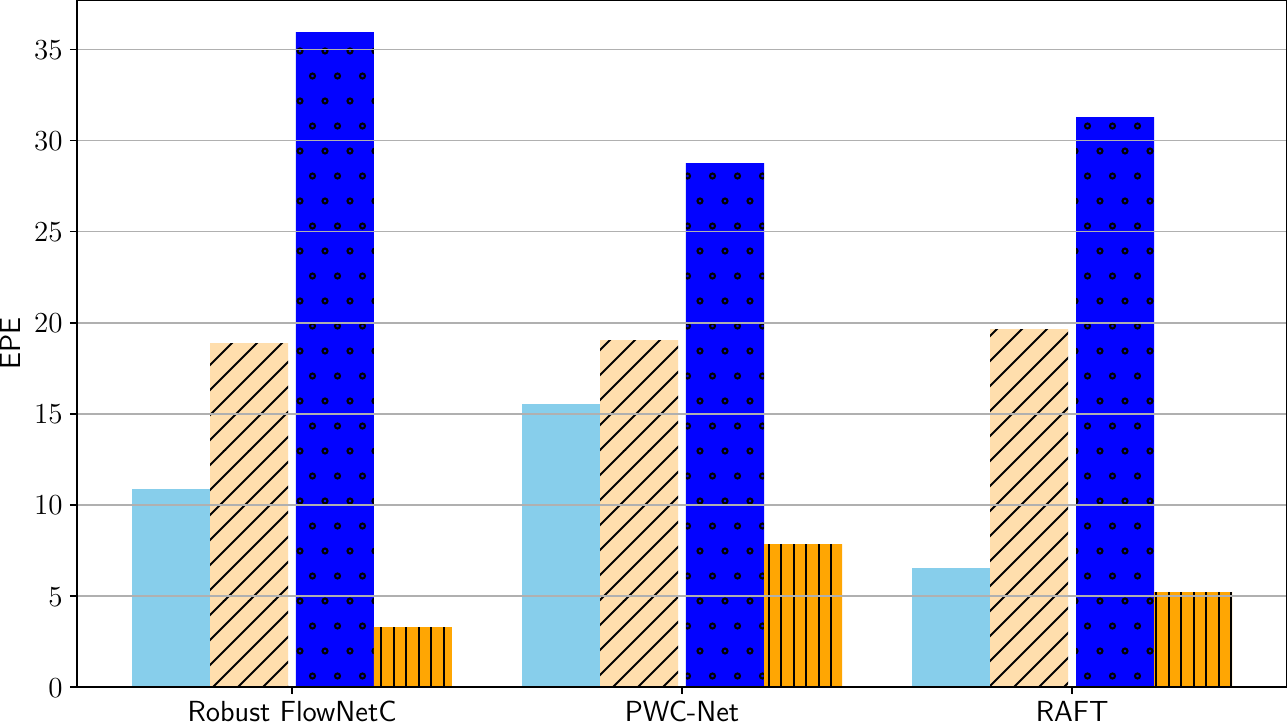}
        \caption{$\epsilon=0.01$}
    \end{subfigure}
    \begin{subfigure}[b]{0.48\textwidth}
        \centering
        \includegraphics[width=\textwidth]{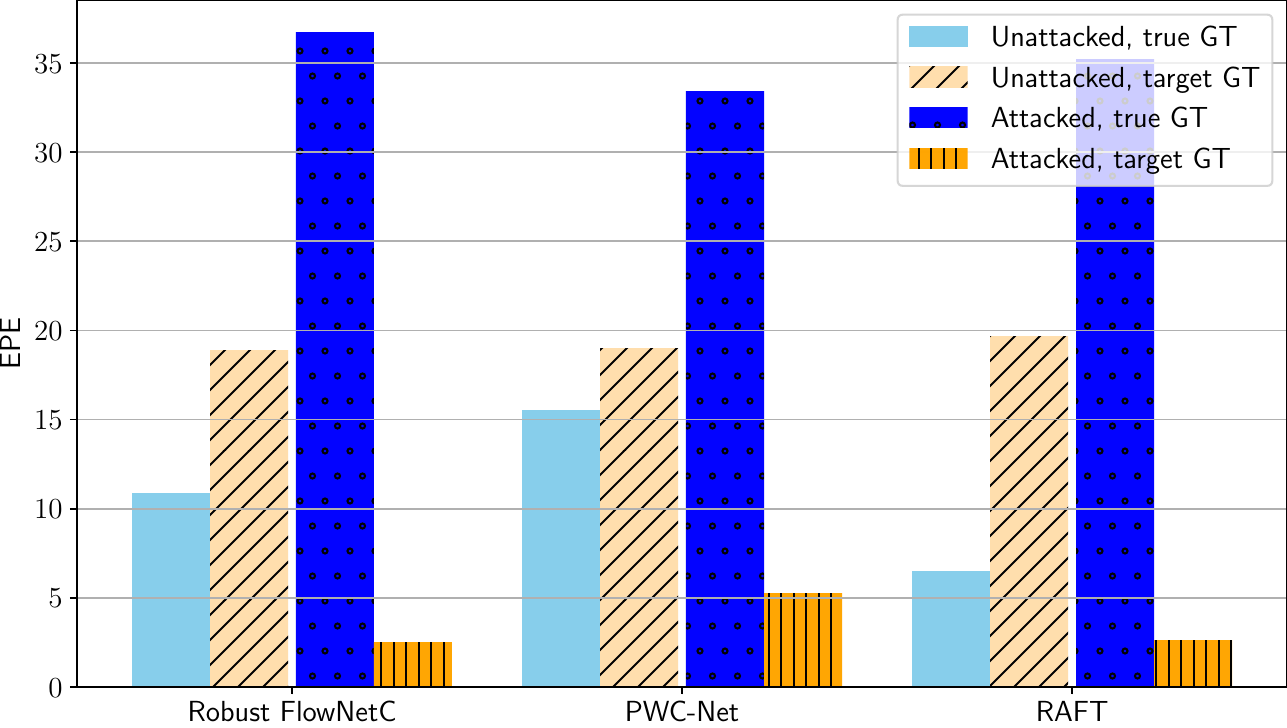}
        \caption{$\epsilon=0.02$}
    \end{subfigure}
    \caption{\textbf{Targeted adversarial attacks for different flow networks and $L_\infty$ norms.} We show the effectiveness of targeted adversarial attacks across different flow networks and $L_\infty$ norms (\ie, $\epsilon=\lbrace 0.002, 0.005, 0.01, 0.02\rbrace$). We computed EPE for each of the unattacked and attacked image frames for both the true ground truth (true GT) and adversarial target ground truth (target GT).
    Across all $L_\infty$ norms the flow estimates get closer to the adversarial target flow. The larger the $L_\infty$ norm (\eg., $\epsilon=0.02$), the larger the shift.
    }
    \label{fig:targetedResults}
\end{figure*}
Figure~\ref{fig:targetedExamples} shows additional examples for targeted adversarial attacks on optical flow networks. The flow estimates are closer to the adversarial target flow than the true flow. 
To quantify our results, also for different $L_\infty$ norms (\ie, $\epsilon=\lbrace 0.002, 0.005, 0.01, 0.02\rbrace$), we ran targeted adversarial attacks for different image pairs from the KITTI 2015 training dataset. Due to computational reasons, we randomly picked a subset of $10$ image pairs. 
Figure~\ref{fig:targetedResults} shows that the resulting flow is closer to the adversarial target flow than the true flow. Note that the resulting flow is closer to the adversarial target flow when the $L_\infty$ norm is larger.

\section{Examples for Adversarial Universal Attacks}\label{sec:universalAttacks}
\begin{figure*}
    \centering
    \begin{subfigure}[b]{0.49\textwidth}
        \centering
        \includegraphics[width=\textwidth]{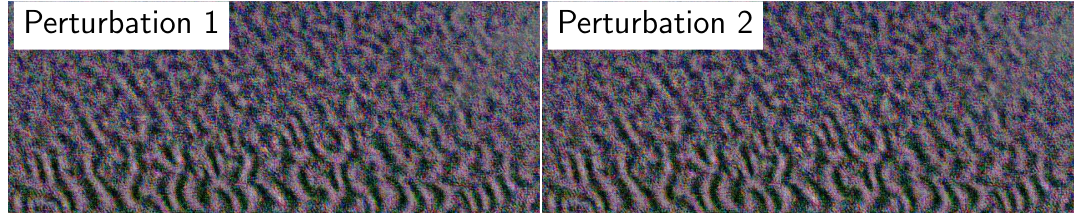}
        \caption{Robust FlowNetC.}
    \end{subfigure}
    \begin{subfigure}[b]{0.49\textwidth}
        \centering
        \includegraphics[width=\textwidth]{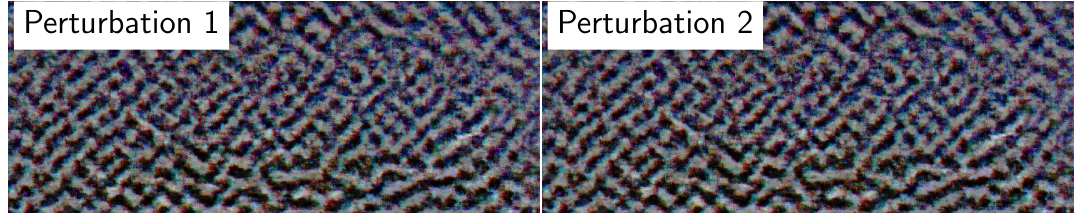}
        \caption{PWC-Net~\cite{sun2018pwc}.}
    \end{subfigure}
    \hfill
    \begin{subfigure}[b]{0.49\textwidth}
        \centering
        \includegraphics[width=\textwidth]{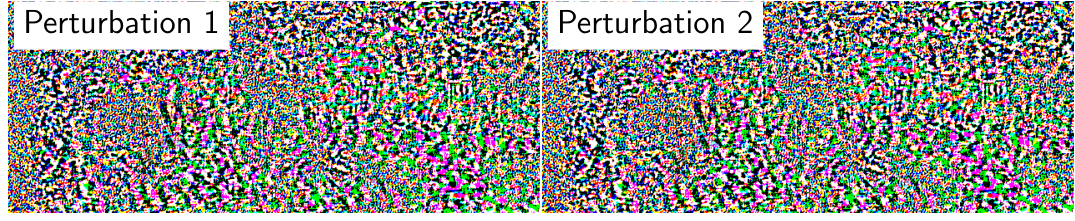}
        \caption{RAFT~\cite{teed2020raft}.}
    \end{subfigure}
    \caption{\textbf{Adversarial universal perturbations.} We show the best found adversarial universal perturbations with $L_\infty$ norm $\epsilon=0.02$ for each flow network for the first and second image frame left or right, respectively. Note that for both Robust FlowNetC and PWC-Net the adversarial universal perturbations contain well-visible self-similar patterns.
    Best viewed in color and with zoom.}
    \label{fig:universalPerturbations}
\end{figure*}
\begin{figure*}
    \centering
    \begin{subfigure}[b]{0.49\textwidth}
        \centering
        \includegraphics[width=\textwidth]{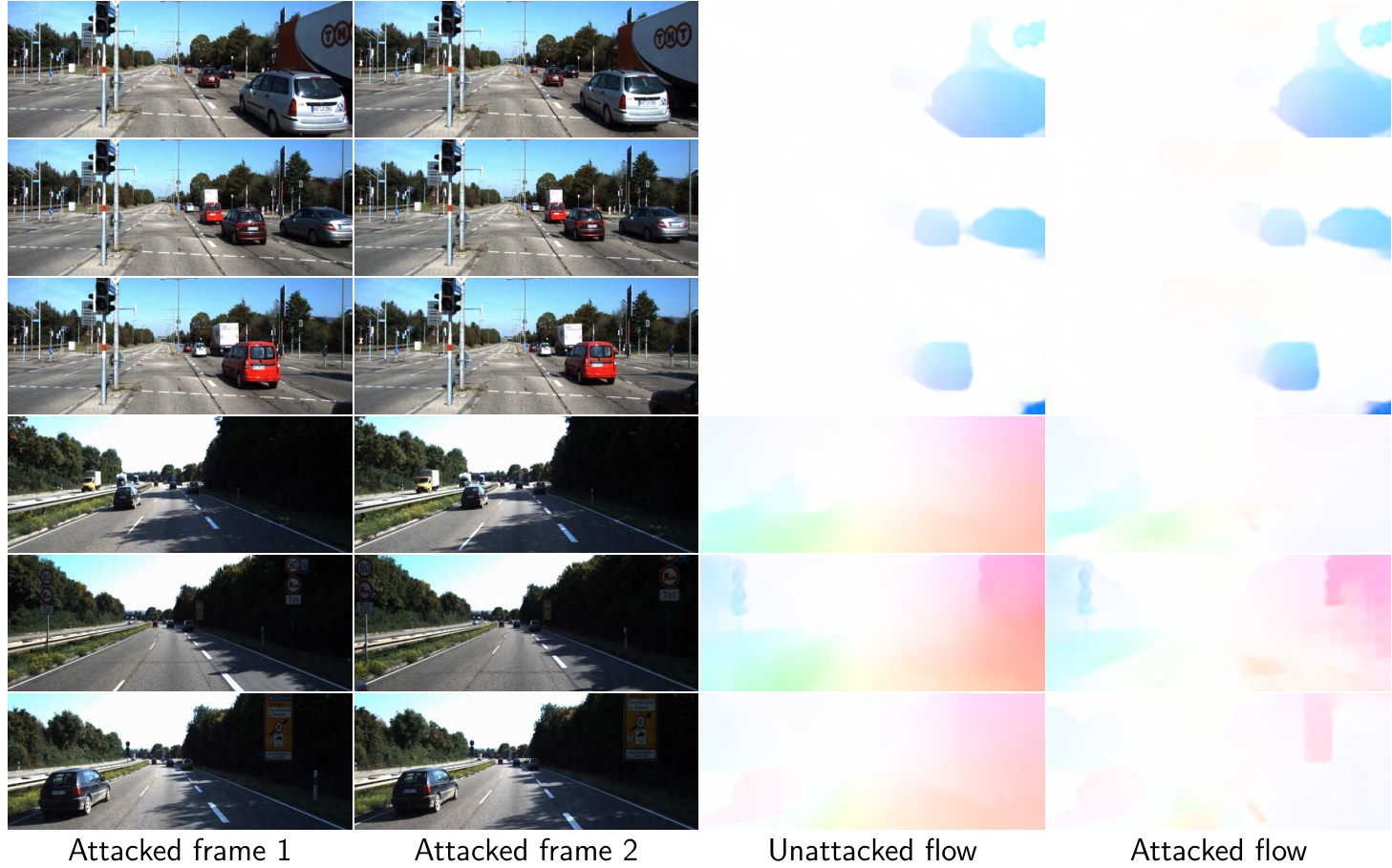}
        \caption{Robust FlowNetC.}
    \end{subfigure}
    \begin{subfigure}[b]{0.49\textwidth}
        \centering
        \includegraphics[width=\textwidth]{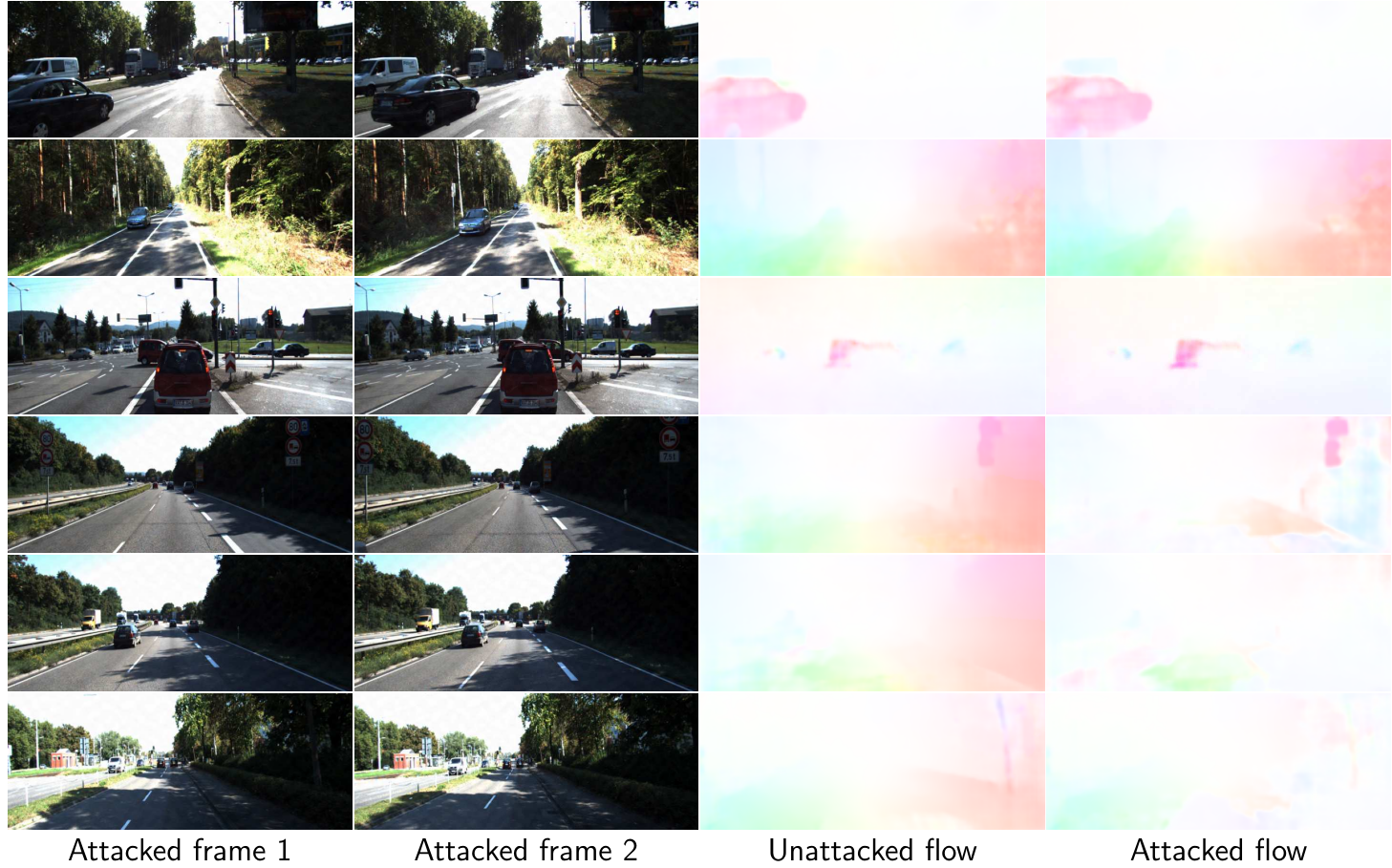}
        \caption{PWC-Net~\cite{sun2018pwc}.}
    \end{subfigure}
    \hfill
    \begin{subfigure}[b]{0.49\textwidth}
        \centering
        \includegraphics[width=\textwidth]{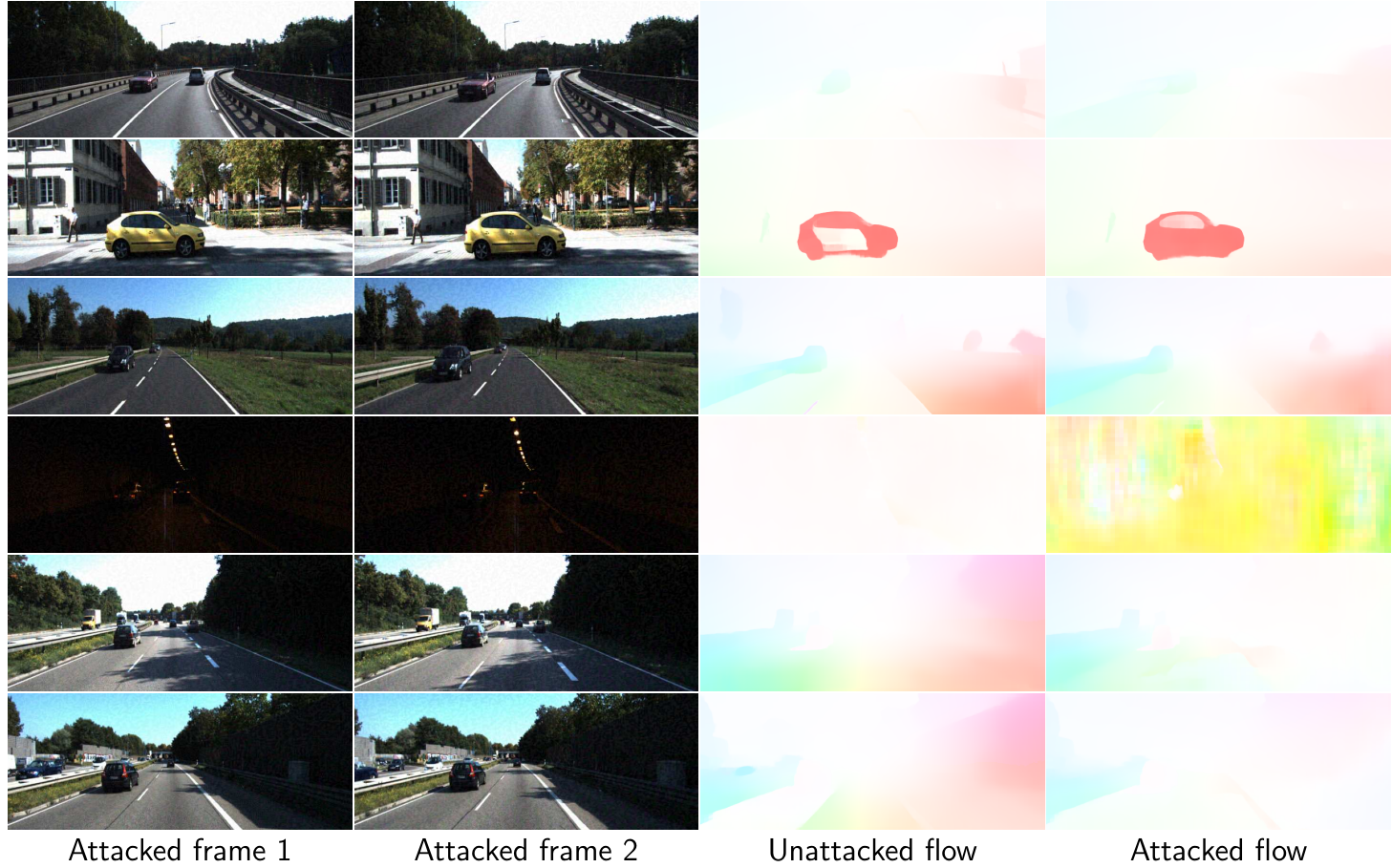}
        \caption{RAFT~\cite{teed2020raft}.}
    \end{subfigure}
    \caption{\textbf{Adversarial universal perturbation attack examples.} Adversarial universal perturbation attacks with I-FGSM and $L_\infty$ norm $\epsilon=0.02$. The bottom three rows of each subfigure show the worst examples based on the absolute degradation of the flow estimate in descending order. Note that the deterioration of flow estimates is to be expected for these image pairs since there are more ambiguities due to the lower contrast. Best viewed in color and with zoom.}
    \label{fig:universalPerturbationsExamples}
\end{figure*}
Figure~\ref{fig:universalPerturbations} shows universal perturbations for the $L_\infty$ norm $\epsilon =0.02$. Note that we can observe well-visible self-similar patterns for Robust FlowNetC and PWC-Net.
Figure~\ref{fig:universalPerturbationsExamples} shows examples for universal attacks with $L_\infty$ norm $\epsilon =0.02$. The flow networks are largely unaffected by the universal perturbations. However, there are some worst-case examples: if there are darker, homogeneous areas, \eg, shadows, in the image frames (and/or there is large ego-motion), the flow deteriorates more. However, this is to be expected because the lower contrast (and large ego-motion) make the estimation problem more difficult, leading to more ambiguities. 
An attacker could exploit this by overwriting the true flow with the help of adversarial ambiguities.

\section{Adversarial Data Augmentation}\label{sec:adv_augmentation}
\begin{figure*}
    \centering
    \begin{subfigure}[b]{0.33\textwidth}
        \centering
        \includegraphics[width=\textwidth]{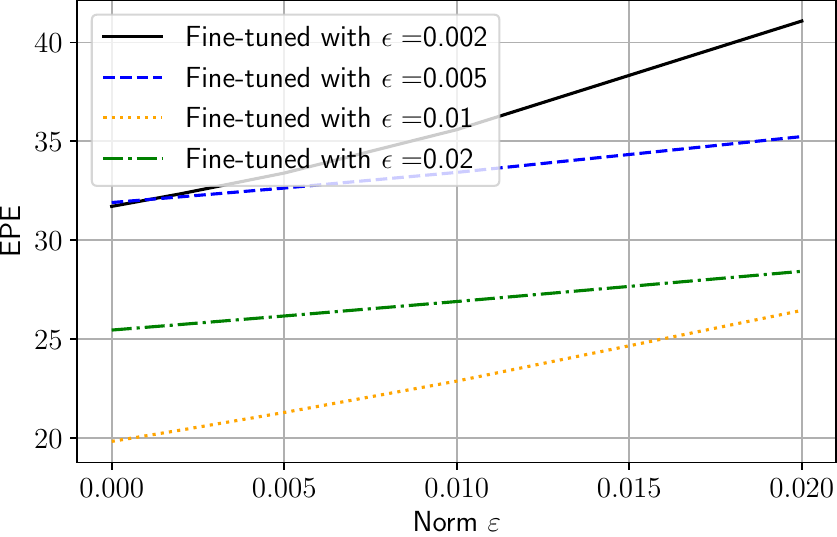}
        \caption{Robust FlowNetC vs. I-FGSM.}
    \end{subfigure}
    \begin{subfigure}[b]{0.33\textwidth}
        \centering
        \includegraphics[width=\textwidth]{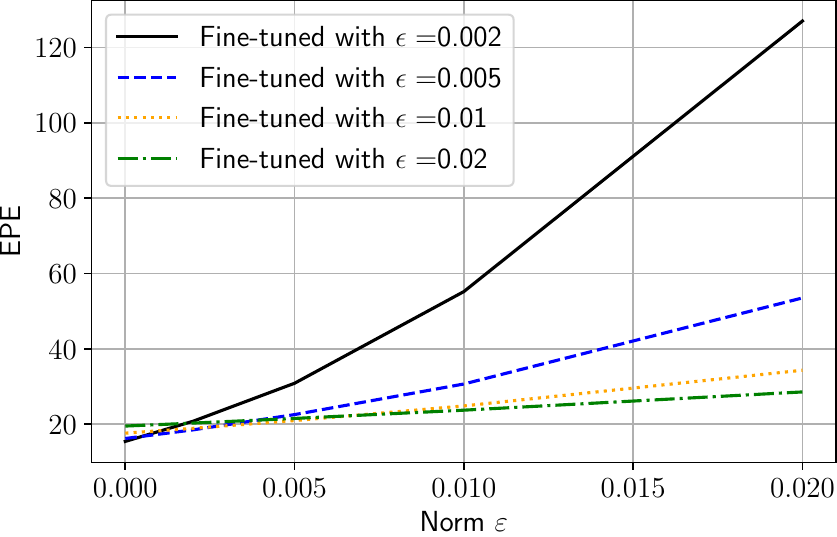}
        \caption{PWC-Net~\cite{sun2018pwc} vs. I-FGSM.}
    \end{subfigure}
    \hfill
    \begin{subfigure}[b]{0.33\textwidth}
        \centering
        \includegraphics[width=\textwidth]{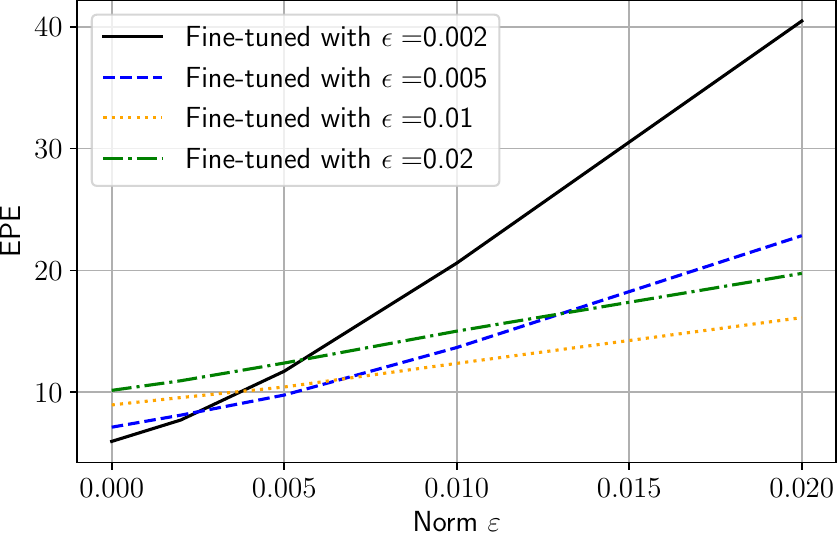}
        \caption{RAFT~\cite{teed2020raft} vs. I-FGSM.}
    \end{subfigure}
    \begin{subfigure}[b]{0.33\textwidth}
        \centering
        \includegraphics[width=\textwidth]{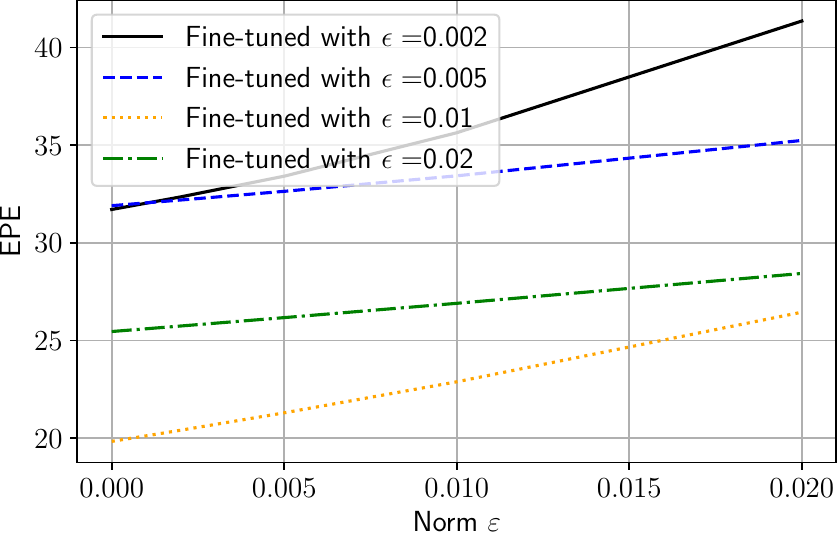}
        \caption{Robust FlowNetC vs. MI-FGSM.}
    \end{subfigure}
    \begin{subfigure}[b]{0.33\textwidth}
        \centering
        \includegraphics[width=\textwidth]{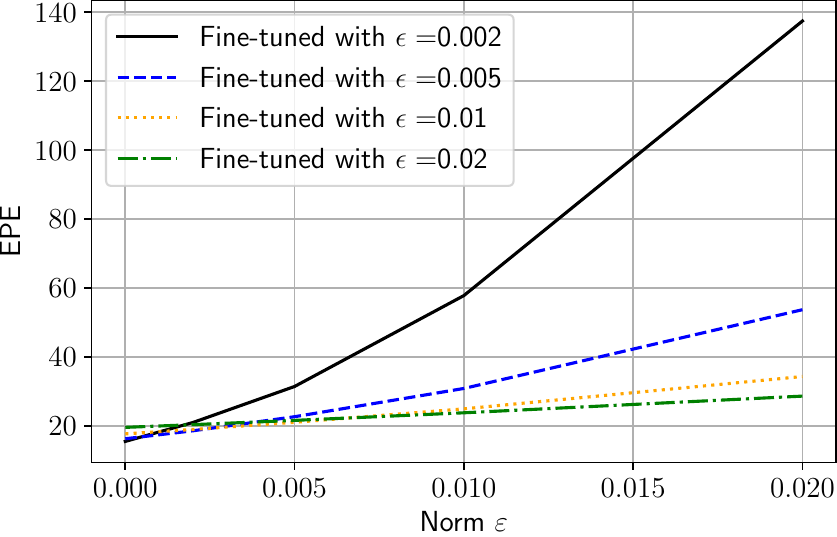}
        \caption{PWC-Net~\cite{sun2018pwc} vs. MI-FGSM.}
    \end{subfigure}
    \hfill
    \begin{subfigure}[b]{0.33\textwidth}
        \centering
        \includegraphics[width=\textwidth]{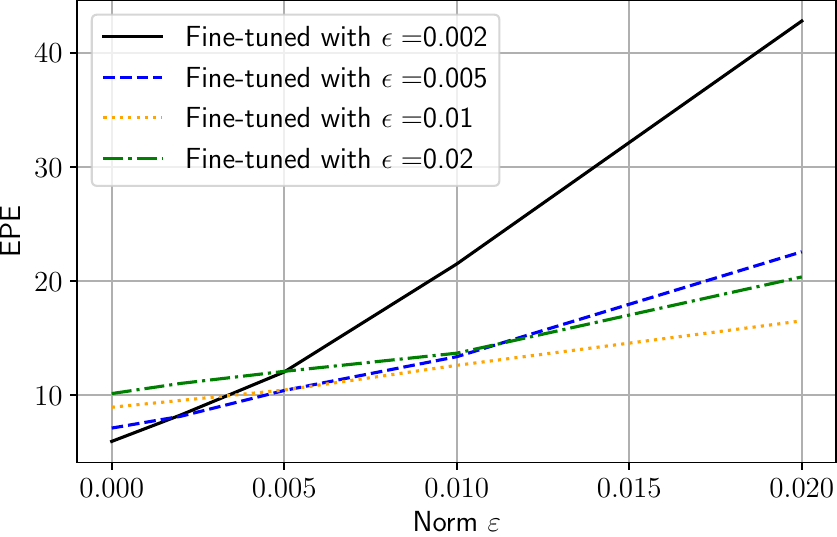}
        \caption{RAFT~\cite{teed2020raft} vs. MI-FGSM.}
    \end{subfigure}
    \caption{\textbf{Adversarial data augmentation.}
    Fine-tuning makes the flow networks more robust \wrt (untargeted) adversarial attacks. We show fine-tuned versions of Robust FlowNetC, PWC-Net and RAFT using various $L_\infty$ norms (\ie, $\epsilon=\lbrace 0.002, 0.005, 0.01, 0.02\rbrace$). For PWC-Net and RAFT, fine-tuning significantly improves robustness \wrt adversarial attacks, while having only a minor negative effect on flow performance for unattacked images. For Robust FlowNetC, flow performance deteriorates significantly for unattacked images. Surprisingly, all flow networks are also robust against the stronger MI-FGSM attacks, although they were not fine-tuned for them.
    }
    \label{fig:adversarialDataAugmentation}
\end{figure*}
Wong~\etal~\cite{wong2021stereopagnosia} showed that they could increase robustness with little negative effect on performance through adversarial data augmentation. To do this, they crafted adversarial examples using FGSM before the adversarial training and added them to the training set.

Different from Wong~\etal, we did not pre-compute adversarial examples but computed them during the training (as typically done in adversarial training for recognition networks). We crafted adversarial examples with the I-FGSM attack (with various $L_\infty$ norms $\epsilon=\lbrace 0.002, 0.005, 0.01, 0.02\rbrace$). We set hyperparameters for the untargeted adversarial attack, as described in Supplement Section~\ref{sec:evalDetails}. We used all $194$ image pairs of the KITTI 2012 test dataset for adversarial training and resized images to $256\times 640$. We chose learning rates $0.000125$ for RAFT, and $0.00001$ for Robust FlowNetC and PWC-Net, and weight decays $0.0001$ for all flow networks. We fine-tuned the flow networks for $30000$ steps with batch size $2$ (\ie, the unattacked and attacked image frames). We did not apply any other data augmentation to the images. For evaluation, we crafted new, unseen (untargeted) adversarial examples on the KITTI 2015 training dataset, as described in Supplement Section~\ref{sec:evalDetails}. In addition, we attacked with the MI-FGSM attack~\cite{dong2018boosting} to evaluate the robustness of flow networks against a stronger, unseen adversarial attack.

Figure~\ref{fig:adversarialDataAugmentation} shows that fine-tuning with adversarial data augmentation improves the robustness of all flow networks. Surprisingly, the adversarial trained flow networks are also robust against the stronger MI-FGSM attacks. Similar to Wong~\etal, we find that contrary to findings in classification~\cite{kurakin2016adversarial}, training with adversarial data augmentation has little negative effect on the performance of optical flow networks (except for Robust FlowNetC). For example, for RAFT, EPE only deteriorates from $5.47$ to $5.96$, $7.13$, $8.96$ and $10.15$ for $L_\infty$ norms $0.002$, $0.005$, $0.01$ and $0.02$ on unattacked image frames, respectively. In general, the smaller the norm the less drop in EPE on unattacked image frames. On the contrary, however, the larger the $L_\infty$ norm, the higher the robustness against adversarial attacks. However, unlike Wong~\etal, we found that training on smaller $L_\infty$ norms (\eg, $\epsilon =0.002$) cannot (significantly) improve robustness on large $L_\infty$ norms (\eg, $\epsilon =0.02$).
We leave further analysis for future work.

\section{Common Image Corruptions}
\begin{figure*}
    \centering
    \begin{subfigure}[b]{0.38\textwidth}
        \centering
        \includegraphics[width=\textwidth]{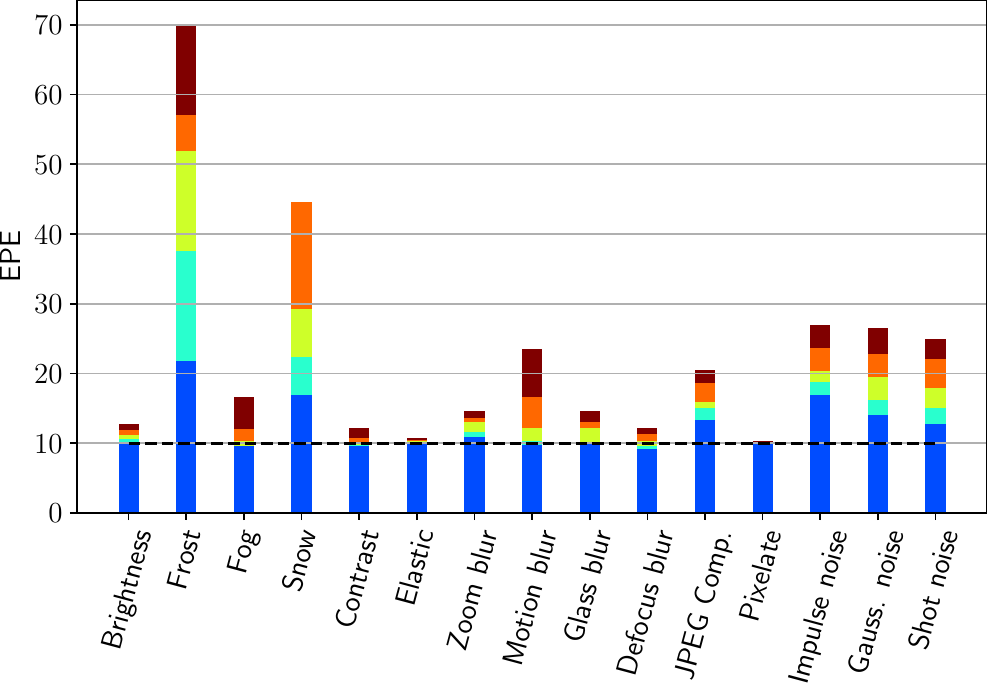}
        \caption{Robust FlowNetC.}
    \end{subfigure}
    \hspace{2cm}
    \begin{subfigure}[b]{0.38\textwidth}
        \centering
        \includegraphics[width=\textwidth]{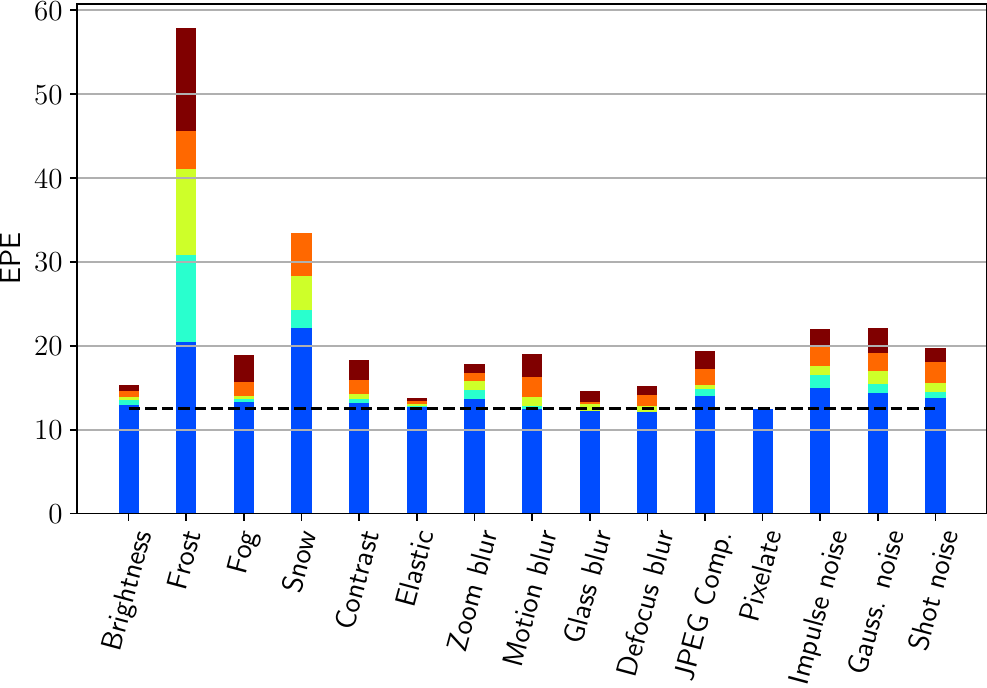}
        \caption{PWC-Net~\cite{sun2018pwc}.}
    \end{subfigure}
    \begin{subfigure}[b]{0.48\textwidth}
        \centering
        \includegraphics[width=\textwidth]{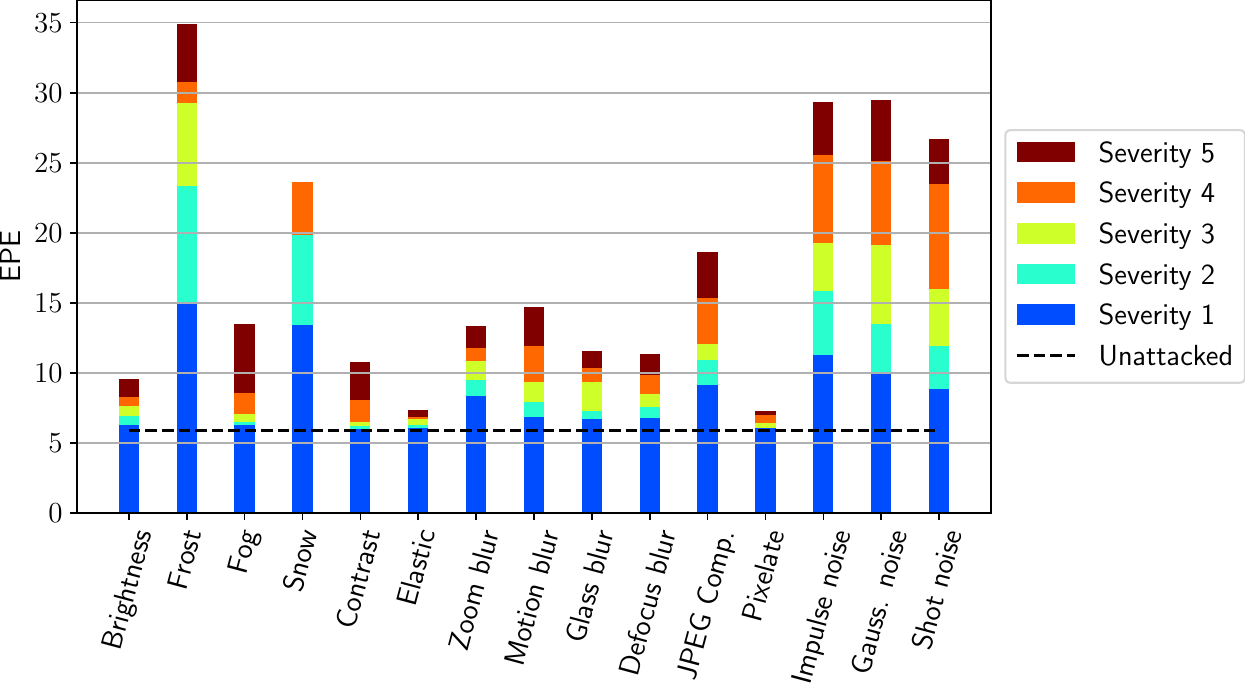}
        \caption{RAFT~\cite{teed2020raft}.}
    \end{subfigure}
    \caption{\textbf{Common image corruptions.} X-axis: corruption types ordered from low to high frequency (based on Saikia~\etal~\cite{SB21b}). Y-axis: EPE for a given corruption type across various severities. All corruption types, except for Frost, Snow, Impulse noise, Gaussian noise, and Shot noise, have little negative effect on flow performance.
    }
    \label{fig:commonCorruptions}
\end{figure*}
In this paper, we focused on adversarial attacks. However, (white-box) adversarial attacks are difficult to apply in the real world, and common image corruptions~\cite{hendrycks2019robustness}, \eg, snow, are more likely to occur. Thus, for completeness, we also studied the robustness of flow networks against common image corruptions across various severities~\cite{michaelis2019dragon}. Similar to our patch-based experiments, we also resized images to $384\times 1280$. Figure~\ref{fig:commonCorruptions} shows that all flow networks are robust against most common image corruptions. However, there are corruptions (\eg, Frost, Snow, Impulse noise, Gaussian noise, Shot noise) that can cause severe deterioration of flow estimates. We suspect that this deterioration is due to some part to the superposition of another flow, \eg, snowfall. We leave further analysis for future work.